\documentclass[10pt,twocolumn,letterpaper]{article}

\usepackage{caption} 
\usepackage{times}
\usepackage{epsfig}
\usepackage{graphicx}
\usepackage{amsmath}
\usepackage{amssymb}
\usepackage{booktabs}
\usepackage{color}
\usepackage{xcolor}
\usepackage{placeins}
\usepackage{iccv}

\iccvfinalcopy 

\usepackage[pagebackref=true,breaklinks=true,letterpaper=true,colorlinks,bookmarks=false,hypertexnames=false]{hyperref}

\usepackage[capitalize]{cleveref}
\crefname{section}{Sec.}{Secs.}
\Crefname{section}{Section}{Sections}
\Crefname{table}{Table}{Tables}
\crefname{table}{Tab.}{Tabs.}

\def\iccvPaperID{6199} 

\begin{document}
\def \modelname {\mbox{BeLFusion}}
\def \participants {126}
\def \supp {supp. material} 
\renewcommand\figureautorefname{Fig.}
\renewcommand\equationautorefname{Eq.}
\renewcommand\tableautorefname{Tab.}
\renewcommand\subsectionautorefname{Sec.}
\renewcommand\sectionautorefname{Sec.}
\newcommand\myeq{\mkern1.5mu{=}\mkern1.5mu}
\def \tochange {red}
\def \changed {blue}
\def \changedtwo {purple}

\title{\modelname{:} Latent Diffusion for Behavior-Driven Human Motion Prediction}

\author{German Barquero\hspace{0.7cm} Sergio Escalera\hspace{0.7cm} Cristina Palmero \\
Universitat de Barcelona and Computer Vision Center, Spain \\
{\tt\small \{germanbarquero, sescalera\}@ub.edu}, {\tt\small crpalmec7@alumnes.ub.edu} \\
\small\url{https://barquerogerman.github.io/BeLFusion/}
}
\twocolumn[{%
\renewcommand\twocolumn[1][]{#1}%
\maketitle
\begin{center}
    \vspace{-0.5cm}
    \captionsetup{type=figure}
    \includegraphics[width=\textwidth,height=5cm]{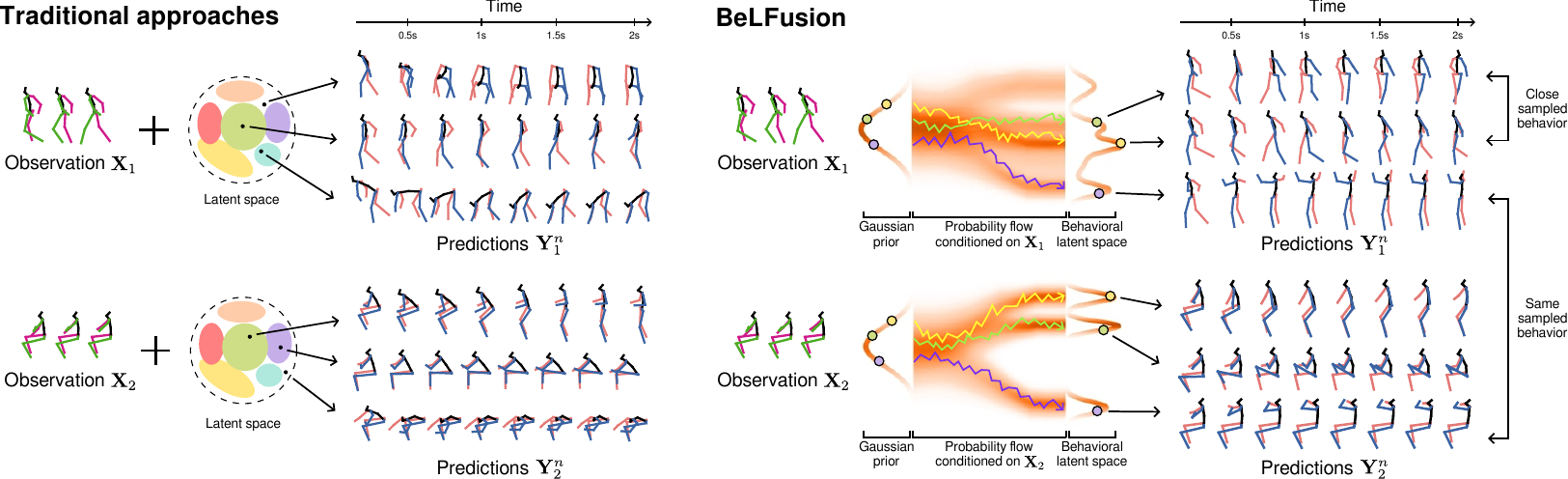}
    \vspace{-0.3cm}
    \captionof{figure}{Common approaches for stochastic human motion prediction use variational autoencoders to model a latent space. Then, the latent code sampled from it is fed to a decoder conditioned on the observation to generate the prediction. In this scenario, out-of-distribution samples or low KL regularizations lead to unrealistic generated sequences. For example, the first prediction for $\mathbf{X}_1$ shows an abrupt and unrealistic transition from walking to bending down. Instead, \modelname{} leverages latent diffusion models to conditionally sample from a behavioral space. Then, samples codes are decoded into predictions that coherently and smoothly transition into a wide range of behaviors.}
    \label{fig:intro}
    \vspace{0.1cm} 
\end{center}%
}]

\ificcvfinal\thispagestyle{empty}\fi

\begin{abstract}
    \vspace{-0.3cm}
    Stochastic human motion prediction (HMP) has generally been tackled with generative adversarial networks and variational autoencoders. Most prior works aim at predicting highly diverse motion in terms of the skeleton joints' dispersion. This has led to methods predicting fast and divergent movements, which are often unrealistic and incoherent with past motion. Such methods also neglect scenarios where anticipating diverse short-range behaviors with subtle joint displacements is important.
    To address these issues, we present \modelname{}, a model that, for the first time, leverages latent diffusion models in HMP to sample from a behavioral latent space where behavior is disentangled from pose and motion. 
    Thanks to our behavior coupler, which is able to transfer sampled behavior to ongoing motion, \modelname{}'s predictions display a variety of behaviors that are significantly more realistic, and coherent with past motion than the state of the art. 
    To support it, we introduce two metrics, the Area of the Cumulative Motion Distribution, and the Average Pairwise Distance Error, which are correlated to realism according to a qualitative study (\participants{} participants). 
    Finally, we prove \modelname{}'s generalization power in a new cross-dataset scenario for stochastic HMP.
\end{abstract}

\vspace{-1.2cm}
\section{Introduction}
\label{sec:intro}

Humans excel at inattentively predicting others' actions and movements. This is key to effectively engaging in social interactions, driving a car, or walking across a crowd. Replicating this ability is imperative in many applications like assistive robots, virtual avatars, or autonomous cars~\cite{andrist2015look, rudenko2020human}. 
Many prior works conceive Human Motion Prediction (HMP) from a deterministic point of view, forecasting a single sequence of body poses, or \textit{motion}, given past poses, usually represented with skeleton joints~\cite{lyu20223d}. 
However, humans are spontaneous and unpredictable creatures by nature, and this deterministic interpretation does not fit contexts where anticipating all possible outcomes is crucial. Accordingly, recent works have attempted to predict the whole distribution of possible future motions (i.e., a \textit{multimodal} distribution) given a short observed motion sequence. We refer to this reformulation as stochastic HMP.

Most prior stochastic works focus on predicting a highly \textit{diverse} distribution of motions.  
Such diversity has been traditionally defined and evaluated in the coordinate space~\cite{yuan2020dlow, dang2022diverse, mao2021gsps, salzmann2022motron, ma2022multiobjective}. This definition biases research toward models that generate fast transitions into very different poses coordinate-wise (see~\autoref{fig:intro}). 
Although there are scenarios where predicting low-speed diverse motion is important, this is discouraged by prior techniques. For example, in assistive robotics, anticipating \textit{behaviors} (i.e., actions) like whether the interlocutor is about to shake your hand or scratch their head might be crucial for preparing the robot's actuators on time~\cite{Barquero2022survey, palmero2022chalearn}. In a surveillance scenario, a foreseen harmful behavior might not differ much from a well-meaning one when considering only the poses along the motion sequence. We argue that this behavioral perspective is paramount to build next-generation stochastic HMP models.
Moreover, results from prior diversity-centric works~\cite{mao2021gsps, dang2022diverse} often suffer from a trade-off that has been persistently overlooked: predicted motion does not look coherent with respect to the latest observed motion. The strong diversity regularization techniques employed often produce abrupt speed changes or direction discontinuities. We argue that consistency with the immediate past is a requirement for prediction plausibility.

To tackle these issues, we present \modelname{} 
(\autoref{fig:intro}). 
By constructing a latent space that disentangles behavior from poses and motion, diversity is no longer limited to the traditional coordinate-based perspective. Instead, diversity is viewed through a behavioral lens, allowing both short- (e.g., hand-waving or smoking) and long-range motions (e.g., standing up or sitting down) to be equally encouraged and represented in the space.
Our \textit{behavior coupler} ensures the predicted behavior is decoded into a smooth and plausible continuation of any ongoing motion. Thus, our predicted motions look more realistic and coherent with the near past than alternatives, which we assess through quantitative and qualitative analyses. In addition, \modelname{} is the first approach that exploits conditional latent diffusion models (LDM)~\cite{vahdat2021score, rombach2022high} for stochastic HMP, achieving state-of-the-art performance. By combining the exceptional capabilities of LDMs to model conditional distributions with the convenient inductive biases of recurrent neural networks (RNNs) for motion modeling~\cite{lyu20223d}, \modelname{} represents a powerful method for stochastic HMP.

To summarize, our main contributions are: (1) We propose \modelname{}, a method that generates predictions that are significantly more realistic and coherent with the near past than prior works, while achieving state-of-the-art accuracy on Human 3.6M~\cite{ionescu2013h36m} and AMASS~\cite{mahmood2019amass} datasets.
(2) We improve and extend the usual evaluation pipeline for stochastic HMP. For the first time in this task, a \textit{cross-dataset evaluation} is conducted to assess the robustness against domain shifts, where the superior generalization capabilities of our method are clearly depicted. This setup, built with AMASS~\cite{mahmood2019amass} dataset, showcases a broad range of actions performed by more than 400 subjects.
(3) We propose two new metrics that provide complementary insights on the statistical similarities between a) the predicted and the dataset averaged absolute motion, and b) the predicted and the intrinsic dataset diversity. We show that they are significantly correlated to our definition of realism.

\def\obsT{B}
\def\predT{T}
\def\xmotion{\mathbf{x}_{m}}
\def\obs{\mathbf{X}}
\def\pred{\mathbf{Y}}

\def\encoder{\mathcal{E}}
\def\decoder{\mathcal{D}}
\def\ldFunction{f_{\Phi}}
\def\latcode{z}
\def\diffused{\latcode_{t}}
\def\diffusedPrev{\latcode_{t-1}}
\def\diffusedStart{\latcode_{0}}

\def\extPred{\pred_{e}}
\def\bvaeEncParams{\theta}
\def\bvaeDecParams{\phi}
\def\bvaeAuxDecParams{\omega}
\def\bvaeXmotionEncParams{\alpha}
\def\bvaeXmotionEnc{g_{\bvaeXmotionEncParams}}
\def\vaeObsEncParams{\lambda}
\def\vaeObsEnc{h_{\vaeObsEncParams}}
\def\bvaeDec{\mathcal{B}_{\bvaeDecParams}}
\def\bvaeEnc{p_{\bvaeEncParams}}
\def\bvaeAuxDec{r_{\bvaeAuxDecParams}}

\def\lossrec{\mathcal{L}_{rec}}
\def\losslat{\mathcal{L}_{lat}}


\section{Related work}
\label{sec:relatedwork}

\subsection{Human motion prediction}

\textbf{Deterministic scenario.}
Prior works on HMP define the problem as regressing a single future sequence of skeleton joints matching the immediate past, or \textit{observed} motion. This regression is often modeled with RNNs~\cite{fragkiadaki2015recurrent, jain2016structural, martinez2017human, gui2018adversarial, pavllo2018quaternet, liu2019towards} or Transformers~\cite{aksan2021spatio, cai2020learning, martinez2021pose}. Graph Convolutional Networks might be included as intermediate layers to model the dependencies among joints~\cite{li2020dynamic, mao2019learning, dang2021msr, li2021skeleton}. Some methods leverage Temporal Convolutional Networks~\cite{li2018convolutional, medjaouri2022hr} or a simple Multi-Layer Perceptron~\cite{guo2022mlp} to predict fixed-size sequences, achieving high performance. Recently, some works claimed the benefits of modeling sequences in the frequency space \cite{cai2020learning, mao2019learning, mao2020history}. However, none of these solutions can model multimodal distributions of future motions. 

\begin{figure*}[t!]
    \centering
    \includegraphics[width=16.5cm]{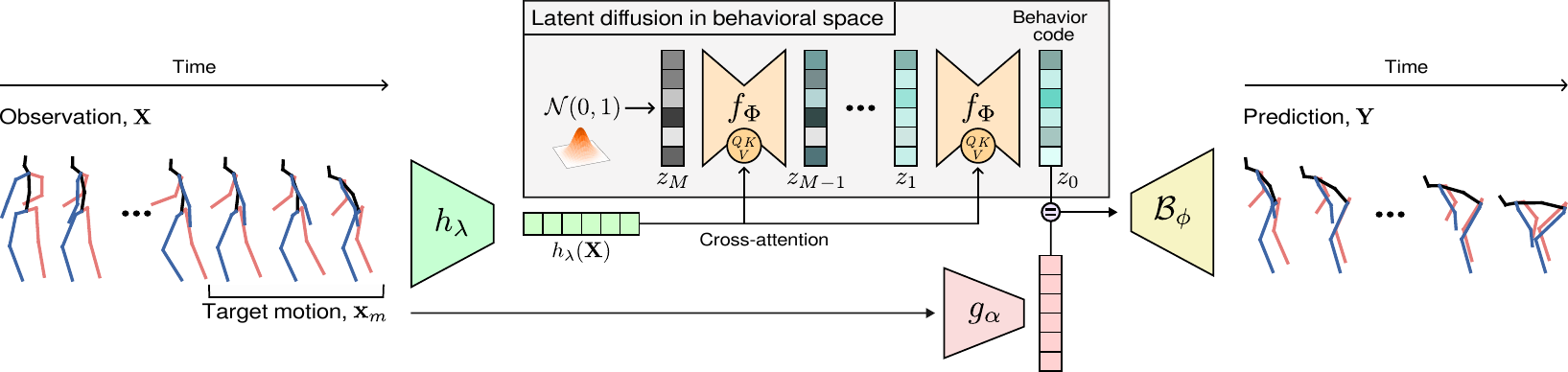}
    \caption{\modelname{}'s architecture. A latent diffusion model conditioned on an encoding of the observation, $\vaeObsEnc(\obs)$, progressively denoises a sample from a zero-mean unit variance multivariate normal distribution into a behavior code. Then, the behavior coupler $\bvaeDec$ decodes the prediction by transferring the sampled behavior to the target motion, $\xmotion$. In our implementation, $\ldFunction$ is a conditional U-Net with cross-attention, $\bvaeXmotionEnc$ is a dense layer, and $\vaeObsEnc$, and $\bvaeDec$ are one-layer recurrent neural networks.}
    \label{fig:main_arch}
    \vspace{-0.2cm}
\end{figure*}

\textbf{Stochastic scenario.} 
To fill this gap, other methods that predict multiple futures for each observed sequence were proposed. Most of them use a generative approach to model the distribution of possible futures. Most popular generative models for HMP are generative adversarial networks (GANs) \cite{barsoum2018hpgan, kundu2019bihmpgan} and variational autoencoders (VAEs) \cite{walker2017theposeknows, yan2018mtvae, cai2021unified, mao2021gsps}. 
These methods often include diversity-promoting losses in order to predict a high variety of motions \cite{mao2021gsps}, or incorporate explicit techniques for diverse sampling~\cite{yuan2020dlow, dang2022diverse,xu2022diverse}. This diversity is computed with the raw coordinates of the predicted poses. We argue that, as a result, the race for diversity has promoted motions deriving to extremely varied poses very early in the prediction. Most of these predictions are neither realistic nor plausible in the context of the observed motion. 
Also, prior works neglect situations where a diversity of behaviors, which can sometimes be subtle, is important. We address this by \textit{implicitly} encouraging such diversity in a behavioral latent space.

\textbf{Semantic human motion prediction. } 
Few works have attempted to leverage semantically meaningful latent spaces for stochastic HMP~\cite{yan2018mtvae, liu2021aggregated, gu2022learning}. For example, \cite{gu2022learning} exploits disentangled motion representations for each part of the body to control the HMP. 
\cite{yan2018mtvae} adds a sampled latent code to the observed encoding to transform it into a prediction encoding. This inductive bias helps the network disentangle a motion code from the observed poses. However, the strong assumption that a simple arithmetic operation can map both sequences limits the expressiveness of the model. Although not specifically focused on HMP, \cite{blattmann2021behavior} proposes an adversarial framework to disentangle a behavioral encoding from a sequence of poses. The extracted behavior can then be transferred to any initial pose. In this paper, we propose a generalization of such framework to transfer behavior to ongoing movements. \modelname{} exploits this disentanglement to improve the behavioral coverage of HMP.

\subsection{Diffusion models}

Denoising diffusion probabilistic models aim at learning to reverse a Markov chain of $M$ diffusion steps (usually $M>100$) that slowly adds random noise to the target data samples~\cite{sohl2015deep, ho2020denoising}. 
For conditional generation, a common strategy consists in applying cross-attention to the conditioning signal at each denoising step~\cite{dhariwal2021diffusionbeatsgans}. 
Diffusion models have achieved impressive results in fields like video generation, inpainting, or anomaly detection~\cite{yang2022diffusionsurvey}. In a more similar context, \cite{rasul2021timegrad, tashiro2021csdi} use diffusion models for time series forecasting. \cite{gu2022stochastic} recently presented a diffusion model for trajectory prediction that controls the prediction uncertainty by shortening the denoising chain. A few concurrent works have explored them for HMP~\cite{saadatnejad2023generic, wei2023human, chen2023humanmac, ahn2023can, chen2023executing}.

However, diffusion models have an expensive trade-off: extremely slow inference due to the large number of denoising steps required. Latent diffusion models (LDM) accelerate the sampling by applying diffusion to a low-resolution latent space learned by a VAE~\cite{vahdat2021score, rombach2022high}. Thanks to the KL regularization, the learned latent space is built close to a normal distribution. As a result, the length of the chain that destroys the latent codes can be greatly reduced, and reversed much faster. 
In this work, we present the first approach that leverages LDM for stochastic HMP, achieving state-of-the-art performance in terms of accuracy and realism.

\def\obsT{B}
\def\predT{T}
\def\xmotion{\mathbf{x}_{m}}
\def\obs{\mathbf{X}}
\def\pred{\mathbf{Y}}

\def\encoder{\mathcal{E}}
\def\decoder{\mathcal{D}}
\def\ldFunction{f_{\Phi}}
\def\latcode{z}
\def\diffused{\latcode_{t}}
\def\diffusedPrev{\latcode_{t-1}}
\def\diffusedStart{\latcode_{0}}

\def\extPred{\pred_{e}}
\def\bvaeEncParams{\theta}
\def\bvaeDecParams{\phi}
\def\bvaeAuxDecParams{\omega}
\def\bvaeXmotionEncParams{\alpha}
\def\bvaeXmotionEnc{g_{\bvaeXmotionEncParams}}
\def\vaeObsEncParams{\lambda}
\def\vaeObsEnc{h_{\vaeObsEncParams}}
\def\bvaeDec{\mathcal{B}_{\bvaeDecParams}}
\def\bvaeEnc{p_{\bvaeEncParams}}
\def\bvaeAuxDec{\mathcal{A}_{\bvaeAuxDecParams}}

\def\lossrec{\mathcal{L}_{rec}}
\def\losslat{\mathcal{L}_{lat}}


\section{Methodology}
\label{sec:methodology}

In this section, we first characterize the HMP problem (\autoref{subsec:definition}). 
Then, we present a straightforward adaptation of conditional LDMs to HMP (\autoref{subsec:molfusion}). Finally, we describe \modelname{}'s keystones (\autoref{fig:main_arch}): our behavioral latent space, the behavioral LDM, and its training losses (\autoref{subsec:behavioral_ld}). 

\subsection{Problem definition}
\label{subsec:definition}

The goal in HMP consists in, given an observed sequence of $\obsT$ poses (\textit{observation window}), predicting the following $\predT$ poses (\textit{prediction window}). In stochastic HMP, $N$ prediction windows are predicted for each observation window.
Accordingly, we define the set of poses in the observation and prediction windows as $\obs{=}\{p_{t-\obsT}, {...}, p_{t-2}, p_{t-1}\}$ and $\pred^{i}{=}\{p_{t}^{i}, p_{t+1}^{i}, {...}, p_{t+\predT-1}^{i}\}$, where $i{\in} \{1, {...}, N\}$\footnote{A sampled prediction $\pred^{i}$ is hereafter referred as $\pred$ for intelligibility. }, and $p_t^{i} {\in} \mathbb{R}^d$ are the coordinates of the human joints at timestep $t$.

\subsection{Motion latent diffusion}
\label{subsec:molfusion}

Here, we define a direct adaptation of LDM to HMP. 
First, a VAE is trained so that an encoder $\encoder$ transforms fixed-length target sequences of $\predT$ poses, $\pred$, into a low-dimensional latent space $V \subset \mathbb{R}^{v}$. Samples $z\in V$ can be drawn and mapped back to the coordinate space with a decoder $\decoder$. 
Then, an LDM conditioned on $\obs$ is trained to predict the corresponding latent vector $\latcode = \encoder(\pred) \in V$\footnote{For simplicity, we use $\encoder(\pred)$ to refer to the expected value of $\encoder(z | \pred)$.}. The generative HMP problem is formulated as follows:
\vspace{-0.15cm}
\begin{equation}
\label{eq:ldm_generative}
P(\pred|\obs)=P(\pred, \latcode|\obs)=P(\pred|\latcode, \obs)P(\latcode|\obs).
\vspace{-0.15cm}
\end{equation}

The first equality holds because $\pred$ is a deterministic mapping from the latent code $\latcode$. Then, sampling from the true conditional distribution $P(\pred|\obs)$ is equivalent to sampling $\latcode$ from $P(\latcode|\obs)$ and decoding $\pred$ with $\decoder$.

LDMs are typically trained to predict the perturbation $\epsilon_t=\ldFunction(\diffused, t, \vaeObsEnc(\obs))$ of the diffused latent code $\diffused$ at each timestep $t$, where $\vaeObsEnc(\obs)$ is the encoded conditioning observation. Once trained, the network $\ldFunction$ can reverse the diffusion Markov chain of length $M$ and infer $\latcode$ from a random sample $\latcode_M \sim \mathcal{N}(0, 1)$. Instead, we choose to use a more convenient parameterization so that $\latcode_0=\ldFunction(\diffused, t, \vaeObsEnc(\obs))$ \cite{xiao2021trilemma, luo2022understandingDM}. With this, an approximation of $\latcode$ is predicted in every denoising step $\diffusedStart$, and used to sample the input of the next denoising step $\diffusedPrev$, by diffusing it $t{-}1$ times. We use $q(\latcode_{t-1} | \latcode_0)$ to refer to this diffusion process. 
With this parameterization, the LDM loss (or \textit{latent} loss) becomes:
\vspace{-0.3cm}
\begin{equation}
\label{eq:lat_loss}
    \losslat(\obs, \pred){=}\sum^T_{t=1}
    \underset{q( \diffused | \diffusedStart  )}{\mathbb{E}}
    \|\ldFunction(\diffused, t, \vaeObsEnc(\obs)) {-} \encoder(\pred)\|_{1}.
    \vspace{-0.3cm}
\end{equation}

Having an approximate prediction at any denoising step allows us to 1) apply regularization in the coordinates space (\autoref{subsec:behavioral_ld}), and 2) stop the inference at any step and still have a meaningful prediction (\autoref{subsec:results}). 


\subsection{Behavioral latent diffusion}
\label{subsec:behavioral_ld}

In HMP, small discontinuities between the last observed pose and the first predicted pose can look unrealistic. Thus, the LDM (\autoref{subsec:molfusion}) must be highly accurate in matching the coordinates of the first predicted pose to the last observed pose. 
An alternative consists in autoencoding the offsets between poses in consecutive frames. Although this strategy minimizes the risk of discontinuities in the first frame, motion speed or direction discontinuities are still bothersome.

Our proposed architecture, \textbf{Be}havioral \textbf{L}atent dif\textbf{Fusion}, or \modelname{}, solves both problems. It reduces the latent space complexity by relegating the adaption of the motion speed and direction to the decoder. It does so by learning a representation of posture-independent human dynamics: a \textit{behavioral representation}. In this framework, the decoder learns to transfer any behavior to an ongoing motion by building a coherent and smooth transition. Here, we first describe how the behavioral latent space is learned, and then detail the \modelname{} pipeline for behavior-driven HMP.

\begin{figure}[t]
    \centering
    \includegraphics[width=0.48\textwidth]{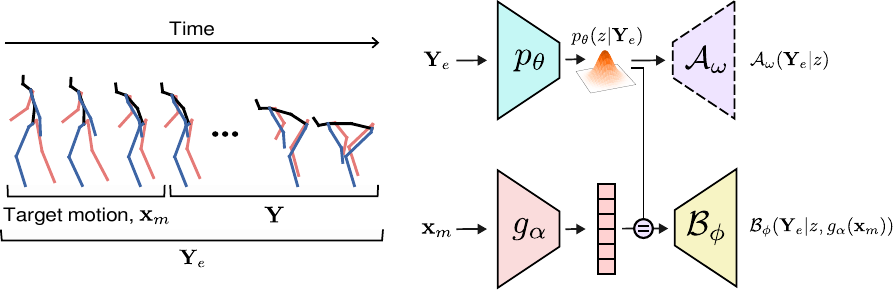}
    \vspace{-0.5cm}
    \caption{Framework for behavioral disentanglement. By adversarially training the auxiliary generator, $\bvaeAuxDec$, against the behavior coupler, $\bvaeDec$, the behavior encoder, $\bvaeEnc$, learns to generate a disentangled latent space of behaviors, $\bvaeEnc(\latcode | \extPred)$. At inference, $\bvaeDec$ decodes a sequence of poses that smoothly transitions from \textit{any} target motion $\xmotion$ to performing the behavior extracted from $\pred$.}
    \label{fig:bvae}
    \vspace{-0.5cm}
\end{figure}

\textbf{Behavioral Latent Space (BLS).} The behavioral representation learning is inspired by \cite{blattmann2021behavior}, which presents a framework to disentangle behavior from motion. Once disentangled, such behavior can be transferred to any static initial pose. We propose an extension of their work to a general and challenging scenario: behavioral transference to ongoing motions. The proposed architecture is shown in \autoref{fig:bvae}.

First, we define the last $C$ observed poses as the \textit{target motion}, $\xmotion=\{p_{t-C}, ..., p_{t-2}, p_{t-1}\}\subset \obs$, and $\extPred=\xmotion \cup \pred$.  $\xmotion$ informs us about the motion speed and direction of the last poses of $\obs$, which should be coherent with $\pred$\footnote{In practice, decoding $\xmotion$ also helped stabilize the BLS training.}. The goal is to disentangle the behavior from the motion and poses in $\extPred$. To do so, we adversarially train two generators, the behavior coupler $\bvaeDec$, and the auxiliary generator $\bvaeAuxDec$, such that a behavior encoder $\bvaeEnc$ learns to generate a disentangled latent space of behaviors $\bvaeEnc (\latcode|\extPred)$. Both $\bvaeDec$ and $\bvaeAuxDec$ have access to such latent space, but $\bvaeDec$ is additionally fed with an encoding of the target motion, $\bvaeXmotionEnc( \xmotion )$. During adversarial training, $\bvaeAuxDec$ aims at preventing $\bvaeEnc$ from encoding pose and motion information by trying to reconstruct poses of $\extPred$ directly from $\bvaeEnc (\latcode|\extPred)$. This training allows $\bvaeDec$ to decode a sequence of poses that smoothly transitions from $\xmotion$ to perform the behavior extracted from $\extPred$. At inference time, $\bvaeAuxDec$ is discarded.

More concretely, the disentanglement is learned by alternating two objectives at each training iteration. The first objective, which optimizes the parameters $\bvaeAuxDecParams$ of the auxiliary generator, forces it to predict $\extPred$ given the latent code $\latcode$:
\vspace{-0.2cm}
\begin{equation}
\label{eq:bvae_aux_loss}
     \underset{\bvaeAuxDecParams}{\max}\; \mathcal{L}_{\text{aux}} = \underset{\bvaeAuxDecParams}{\max}\; \mathbb{E}_{\bvaeEnc(\latcode|\extPred)} (\log \bvaeAuxDec(\extPred |\latcode)).
     \vspace{-0.2cm}
\end{equation}

The second objective acts on the parameters of the target motion encoder, $\bvaeXmotionEncParams$, the behavior encoder, $\bvaeEncParams$, and the behavior coupler, $\bvaeDecParams$. It forces $\bvaeDec$ to learn an accurate $\extPred$ reconstruction through the construction of a normally distributed intermediate latent space:
\vspace{-0.2cm}
\begin{equation}
\label{eq:bvae_main_loss}
\begin{gathered}
    \underset{\bvaeXmotionEncParams, \bvaeEncParams, \bvaeDecParams}{\max}\; \mathcal{L}_{\text{main}} = \underset{\bvaeXmotionEncParams, \bvaeEncParams, \bvaeDecParams}{\max}\; \mathbb{E}_{\bvaeEnc(\latcode|\extPred)} [\log \bvaeDec(\extPred|\latcode, \bvaeXmotionEnc(\xmotion))] \\
    - D_{\text{KL}}(\bvaeEnc(\latcode | \extPred) || p(\latcode))) - \mathcal{L}_{\text{aux}}.
    \vspace{-0.2cm}
\end{gathered}
\end{equation}
Note that the parameters $\bvaeAuxDecParams$ are not optimized when training with \autoref{eq:bvae_main_loss}, and $\bvaeXmotionEncParams, \bvaeEncParams, \bvaeDecParams$ with \autoref{eq:bvae_aux_loss}. The prior $p(\latcode)$ is a multivariate $\mathcal{N}(0, I)$.
The inclusion of $-\mathcal{L}_{\text{aux}}$ in \autoref{eq:bvae_main_loss} penalizes any accurate reconstruction of $\extPred$ through $\bvaeAuxDec$, forcing $\bvaeEnc$ to filter any postural information out. Since $\bvaeDec$ has access to the target posture and motion provided by $\xmotion$, it only needs $\bvaeEnc (\latcode|\extPred)$ to encode the behavioral dynamics. One could argue that a valid alternative strategy for $\bvaeEnc$ would consist in disentangling motion from postures. However, motion dynamics can still be used to extract a good pose approximation.
See \supp{} Sec. C for more details and visual examples of behavioral transference to several motions $\xmotion$.

\textbf{Behavior-driven HMP.} \modelname{}'s goal is to sample the appropriate behavior code given the observation $\obs$, see \autoref{fig:main_arch}.
To that end, a conditional LDM is trained to optimize $\losslat(\obs,\extPred)$ (\autoref{eq:lat_loss}), with
$\encoder = \bvaeEnc$, so that it learns to predict the behavioral encoding of $\extPred$: the expected value of $\bvaeEnc (\latcode|\extPred)$.
Then, the behavior coupler, $\bvaeDec$, transfers the predicted behavior to the target motion, $\xmotion$, to reconstruct the poses of the prediction. 
However, the reconstruction of $\bvaeDec$ is also conditioned on $\xmotion$. Such dependency cannot be modeled by the $\losslat$ objective alone. Thanks to our parameterization (\autoref{subsec:molfusion}), we can also use the traditional MSE loss in the reconstruction space:
\vspace{-0.3cm}
\begin{equation}
\label{eq:rec_loss}
\begin{gathered}
    \lossrec(\obs, \extPred)= 
    \sum^T_{t=1}
    \underset{q(z_t|z_0)}{\mathbb{E}}
    \|\bvaeDec(\ldFunction(z_t, t, \vaeObsEnc(\obs)), \bvaeXmotionEnc(\xmotion)) \\ - \bvaeDec(\bvaeEnc(\extPred), \bvaeXmotionEnc(\xmotion))\|_{2}.
    \vspace{-0.3cm}
\end{gathered}
\end{equation}


The second term of \autoref{eq:rec_loss} is the reconstructed $\extPred$. Optimizing the objective within the solutions space upper bounded by $\bvaeDec$'s reconstruction capabilities helps stabilize the training. Note that only the future poses $\pred \subset \extPred$ form the prediction.
The observation encoder, $\vaeObsEnc$, is pretrained in an autoencoding framework that reconstructs $\obs$. We found experimentally that $\vaeObsEnc$ does not benefit from further training, so its parameters $\vaeObsEncParams$ are frozen when training the LDM. The target motion encoder, $\bvaeXmotionEnc$, and the behavior coupler, $\bvaeDec$, are also pretrained as described before and kept frozen. $\ldFunction$ is conditioned on $\vaeObsEnc(\obs)$ with cross-attention.



\textbf{Implicit diversity loss.} Although training \modelname{} with Eqs.~\ref{eq:lat_loss} and \ref{eq:rec_loss} leads to accurate predictions, their diversity is poor. We argue that this is due to the strong regularization of both losses. Similarly to \cite{fan2017point, gupta2018social}, we propose to relax them by sampling $k$ predictions at each training iteration and only backpropagating the gradients through the two predictions that separately minimize the latent or the reconstructed loss (further discussion in \supp{} Sec.~D.2):
\vspace{-0.2cm}
\begin{equation}
    \label{eq:final_loss}
    \underset{k}{\min}\; \losslat(\obs, \extPred^k) + 
    \lambda \; \underset{k}{\min}\; \lossrec(\obs, \extPred^k),
    \vspace{-0.2cm}
\end{equation}

where $\lambda$ controls the trade-off between the latent and the reconstruction errors. Regularization relaxation usually leads to out-of-distribution predictions~\cite{mao2021gsps}. This is often solved by employing additional complex techniques like pose priors, or bone-length losses that regularize the other predictions~\cite{mao2021gsps,bie2022hitdvae}. \modelname{} can dispense with it due to mainly two reasons: 1) Denoising diffusion models are capable of faithfully capturing a greater breadth of the training distribution than GANs or VAEs~\cite{dhariwal2021diffusionbeatsgans}; 2) The variational training of the behavior coupler makes it more robust to errors in the predicted behavior code.

\section{Experimental evaluation}
\label{sec:experimental_setup}

\setlength{\tabcolsep}{3pt}
\begin{table*}[t!]\renewcommand{\arraystretch}{0.9}
\footnotesize		
\centering
\begin{tabular}{l@{\hskip 3mm}cccccccc@{\hskip 2mm}|@{\hskip 2mm}ccccccc}
\toprule
& \multicolumn{8}{c}{Human3.6M \cite{ionescu2013h36m}} &  \multicolumn{7}{c}{AMASS \cite{mahmood2019amass}} \\

\midrule
 & APD & APDE  & ADE  & FDE   & MMADE  & MMFDE & CMD  & FID* & APD & APDE & ADE  & FDE  & MMADE  & MMFDE & CMD  \\
\midrule
Zero-Velocity & 0.000 & 8.079 & 0.597 & 0.884 & 0.683 & 0.909 & 22.812 & 0.606 & 0.000 & 9.292 & 0.755 & 0.992 & 0.814 & 1.015 & 39.262\\
BeGAN k=1 & 0.675 & 7.411 & 0.494 & 0.729 & 0.605 & 0.769 & 12.082 & 0.542 & 0.717 & 8.595 & 0.643 & 0.834 & 0.688 & 0.843 & 24.483 \\
BeGAN k=5 & 2.759 & 5.335 & 0.495 & 0.697 & 0.584 & 0.718 & 13.973 & 0.578 & 5.643 & 4.043 & 0.631 & 0.788 & 0.667 & 0.787 & 24.034 \\
BeGAN k=50 & 6.230 & 2.200 & 0.470 & 0.637 & 0.561 & 0.661 & 8.406 & 0.569 & 7.234 & 2.548 & 0.613 & 0.717 & 0.650 & 0.720 & 22.625\\
\midrule
HP-GAN~\cite{barsoum2018hpgan} & 7.214 & - & 0.858 & 0.867 & 0.847 & 0.858 & - & - & -  & -  & -  & -  & -  & -  & -  \\
DSF~\cite{yuan2019dsf} & 9.330 & - & 0.493 & 0.592 & 0.550 & 0.599 & - & - & -  & -  & -  & -  & -  & -  & -  \\
DeLiGAN~\cite{gurumurthy2017deligan} & 6.509 & - & 0.483 & 0.534 & 0.520 & 0.545 & - & - & -  & -  & -  & -  & -  & -  & -  \\
GMVAE~\cite{dilokthanakul2016gmvae} & 6.769 & - & 0.461 & 0.555 & 0.524 & 0.566 & - & - & -  & -  & -  & -  & -  & -  & -  \\
TPK~\cite{walker2017theposeknows} & 6.723 & \underline{1.906} & 0.461 & 0.560 & 0.522 & 0.569 & 6.326 & 0.538 & 
9.283 & \underline{2.265} & 0.656 & 0.675 & 0.658 & 0.674 & 17.127\\
MT-VAE~\cite{yan2018mtvae} & 0.403 & - & 0.457 & 0.595 & 0.716 & 0.883 & - & - & -  & -  & -  & -  & -  & -  & -  \\
BoM~\cite{bhattacharyya2018bom} & 6.265 & - & 0.448 & 0.533 & 0.514 & 0.544 & - & - & -  & -  & -  & -  & -  & -  & -  \\
DLow~\cite{yuan2020dlow} & 11.741 & 3.781 & 0.425 & 0.518 & 0.495 & 0.531 & \textbf{4.927} & 1.255 & 
\underline{13.170} & 4.243 & 0.590 & 0.612 & 0.618 & 0.617 & \textbf{15.185}\\
MultiObj~\cite{ma2022multiobjective} & 14.240 & - & 0.414 & 0.516 & - & - & - & - & 
- & - & - & - & - & - & -\\
GSPS~\cite{mao2021gsps} & \underline{14.757} & 6.749 & 0.389 & 0.496 & 0.476 & 0.525 & 10.758 & 2.103 & 
12.465 & 4.678 & 0.563 & 0.613 & 0.609 & 0.633 & 18.404\\
Motron~\cite{salzmann2022motron} & 7.168 & 2.583 & 0.375 & 0.488 & 0.509 & 0.539 & 40.796 & 13.743 & 
- & - & - & - & - & - & -\\
DivSamp~\cite{dang2022diverse} & \textbf{15.310} & 7.479 & \underline{0.370} & 0.485 & 0.475 & 0.516 & 11.692 & 2.083 & 
\textbf{24.724} & 15.837 & 0.564 & 0.647 & 0.623 & 0.667 & 50.239\\
\midrule
BeLFusion\_D & 5.777 & 2.571 & \textbf{0.367} & \textbf{0.472} & \textbf{0.469} & \textbf{0.506} & 8.508 & \underline{0.255} & 7.458 & 2.663 & \textbf{0.508} & \underline{0.567} & \textbf{0.564} & \underline{0.591} & 19.497\\
BeLFusion & 7.602 & \textbf{1.662} & 0.372 & \underline{0.474} & \underline{0.473} & \underline{0.507} & \underline{5.988} & \textbf{0.209} & 9.376 & \textbf{1.977} & \underline{0.513} & \textbf{0.560} & \underline{0.569} & \textbf{0.585} & \underline{16.995}\\
 \bottomrule
    \end{tabular}
    \vspace{-0.1cm}
    \caption{Comparison of \modelname{}\_D (single denoising step) and \modelname{} (all denoising steps) with state-of-the-art methods for stochastic human motion prediction on Human3.6M and AMASS datasets. Bold and underlined results correspond to the best and second-best results, respectively. Lower is better for all metrics except APD. *Only showed for Human3.6M due to lack of class labels for AMASS.}
    \vspace{-0.4cm}
    \label{tab:sota_comparison}
\end{table*}
\setlength{\tabcolsep}{6pt}

Our experimental evaluation is tailored toward two objectives. First, we aim at proving \modelname{}'s generalization capabilities for both seen and unseen scenarios. For the latter, we propose a challenging cross-dataset evaluation setup. Second, we want to demonstrate the superiority of our model with regard to the realism of its predictions compared to state-of-the-art approaches. In this sense, we propose two metrics and perform a qualitative study.

\subsection{Evaluation setup}
\label{subsec:evaluation_setup}

\textbf{Datasets.} 
We evaluate our proposed methodology on Human3.6M~\cite{ionescu2013h36m} (H36M), and AMASS~\cite{mahmood2019amass}. 
H36M consists of clips where 11 subjects perform 15 actions, totaling 3.6M frames recorded at 50 Hz, with action class labels available. We use the splits proposed by \cite{yuan2020dlow} and adopted by most subsequent works~\cite{mao2021gsps, salzmann2022motron, ma2022multiobjective, dang2022diverse} (16 joints). Accordingly, 0.5s (25 frames) are used to predict the following 2s (100 frames). 
AMASS is a large-scale dataset that, as of today, unifies 24 extremely varied datasets with a common joints configuration, with a total of 9M frames when downsampled to 60Hz. Whereas latest deterministic HMP approaches already include a within-dataset AMASS configuration in their evaluation protocol \cite{mao2020history, aksan2021spatio, medjaouri2022hr}, the dataset remains unexplored in the stochastic context yet. 
To determine whether state-of-the-art methods can generalize their learned motion predictive capabilities to other contexts (i.e., other datasets), we propose a new cross-dataset evaluation protocol with AMASS. The training, validation, and test sets include 11, 4, and 7 datasets, and 406, 33, and 54 subjects (21 joints), respectively. We set the observation and prediction windows to 0.5s and 2s (30 and 120 frames after downsampling), respectively. AMASS does not provide action labels. See \supp{} Sec. B for more details. 

\textbf{Baselines.} We include the zero-velocity baseline, which has been proven very competitive in HMP~\cite{martinez2017human, Barquero2022}, and a version of our model that replaces the LDM with a GAN, BeGAN~\cite{gan2014}. We train three versions with $k=1, 5, 50$.
We also compare against state-of-the-art methods for stochastic HMP (referenced in~\autoref{tab:sota_comparison}). For H36M, we took all evaluation values from their respective works. For AMASS, we retrained state-of-the-art methods with publicly available code that showed competitive performance for H36M. 

\textbf{Implementation details.} We trained \modelname{} with $N{=}50$, $M{=}10$, $k{=}50$, a U-Net with cross-attention~\cite{dhariwal2021diffusionbeatsgans} as $\ldFunction$, one-layer RNNs as $\vaeObsEnc$, and $\bvaeDec$, and a dense layer as $\bvaeXmotionEnc$. 
For H36M, $\lambda{=}5$, and for AMASS, $\lambda{=}1$. At inference, we use an exponential moving average of the trained model with a decay of 0.999. Sampling was conducted with a DDIM sampler~\cite{song2021ddim}.
As explained in \autoref{subsec:molfusion}, our implementation of LDM can be early-stopped at any step of the chain of length $M$ and still have access to an approximation of the behavioral latent code.
Thus, we also include \modelname{}'s results when inference is early-stopped right after the first denoising step (i.e., x10 faster): \modelname{}\_D. Further details are included in the \supp{} Sec. A.

\subsection{Evaluation metrics}
\label{subsec:evaluation_metrics} 

To compare \modelname{} with prior works, we follow the well-established evaluation pipeline proposed in \cite{yuan2020dlow}. The Average and the Final Displacement Error metrics (ADE, and FDE, respectively) quantify the error on the most similar prediction compared to the ground truth. While the ADE averages the error along all timesteps, the FDE only does it for the last predicted frame. Their multimodal versions for stochastic HMP, MMADE and MMFDE, compare all predicted futures with the multimodal ground truth of the observation. To obtain the latter, each observation window $\obs$ is grouped with other observations $\obs_i$ with a similar last observed pose in terms of L2 distance. The corresponding prediction windows $\pred_i$ form the \textit{multimodal ground truth} of $\obs$.
The Average Pairwise Distance (APD) quantifies the diversity by computing the L2 distance among all pairs of predicted poses at each timestep. Following \cite{guo2020action2motion, petrovich2021action, dang2022diverse, bie2022hitdvae}, we also include the Fréchet Inception Distance (FID), which leverages the output of the last layer of a pretrained action classifier to quantify the similarity between the distributions of predicted and ground truth motions.

\textbf{Area of the Cumulative Motion Distribution (CMD).} The plausibility and realism of human motion are difficult to assess quantitatively. However, some metrics can provide an intuition of when a set of predicted motions are not plausible. 
For example, consistently predicting high-speed movements given a context where the person was standing still might be plausible but does not represent a statistically coherent distribution of possible futures. We argue that prior works have persistently ignored this. We propose a simple complementary metric: the area under the cumulative motion distribution. First, we compute the average of the L2 distance between the joint coordinates in two consecutive frames (displacement) across the whole test set, $\bar{M}$. Then, for each frame $t$ of all predicted motions, we compute the average displacement $M_{t}$. Then:
\vspace{-0.25cm}
\small\begin{equation}
    \text{CMD} = \sum_{i=1}^{T-1} \sum^i_{t=1} \| M_t - \bar{M}  \|_{1}= \sum_{t=1}^{T-1}(T-t) \| M_t - \bar{M}  \|_{1}.\vspace{-0.15cm}
\end{equation}\normalsize

Our choice to accumulate the distribution is motivated by the fact that early motion irregularities in the predictions impact the quality of the remaining sequence. Intuitively, this metric gives an idea of how the predicted average displacement per frame deviates from the expected one. However, the expected average displacement could arguably differ among actions and datasets. To account for this, we compute the total CMD as the weighted average of the CMD for each H36M action, or each AMASS test dataset, weighted by the action or dataset relative frequency.

\textbf{Average Pairwise Distance Error (APDE).} There are many elements that condition the distribution of future movements and, therefore, 
the appropriate motion diversity levels. To analyze to which extent the diversity is properly modeled, we introduce the average pairwise distance error. We define it as the absolute error between the APD of the multimodal ground truth and the APD of the predicted samples. Samples without any multimodal ground truth are dismissed. See \supp{} Fig. E for a visual illustration.


\begin{figure*}[t!]
    \centering
    \includegraphics[width=\textwidth]{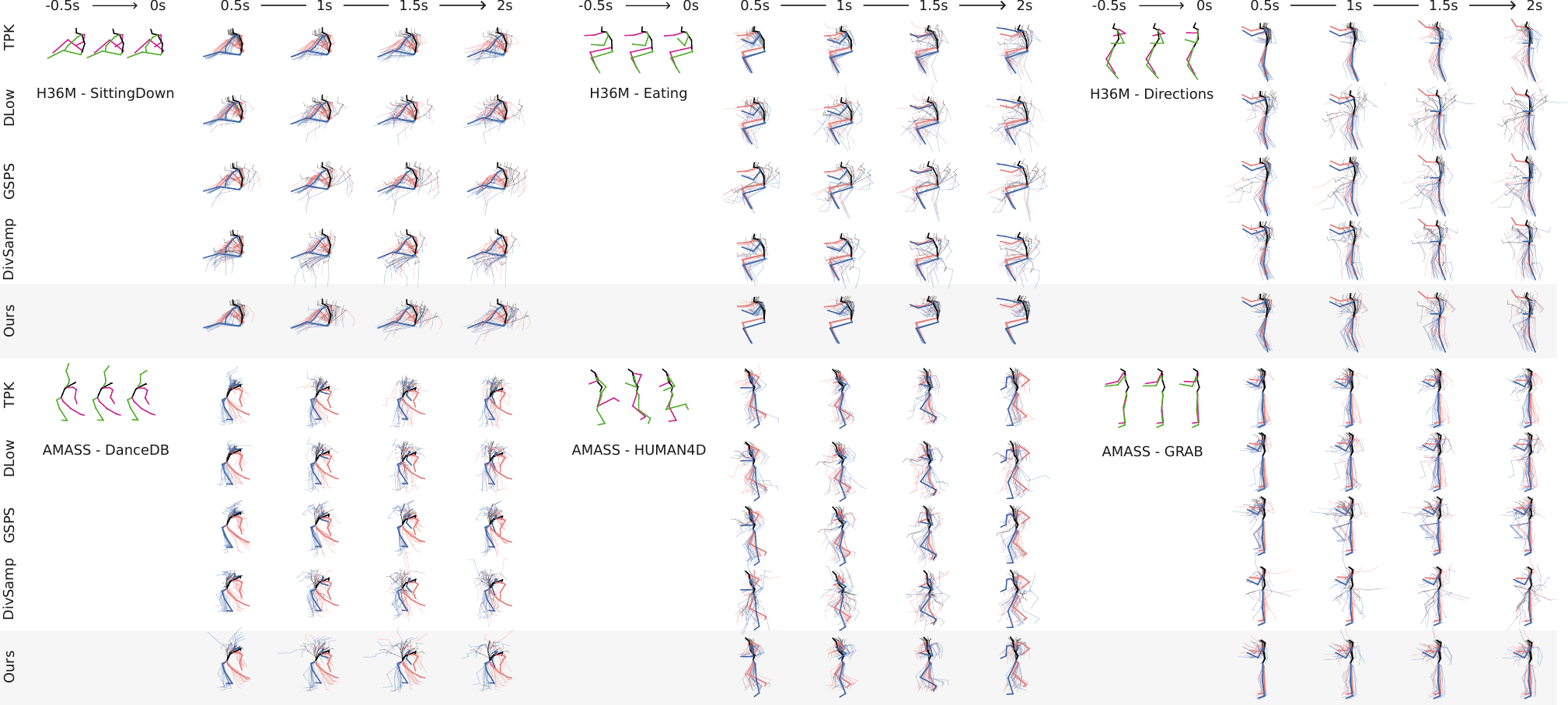}
    \vspace{-0.5cm}
    \caption{Qualitative results show the adaption of \modelname{}'s diversity to the observation context in both within- (H36M, top) and cross-dataset (AMASS, bottom). At each future timestep, 10 predicted samples are superimposed below the thicker ground truth.}
    \label{fig:qualitative_examples}\vspace{-0.4cm}
\end{figure*}

\subsection{Results}
\label{subsec:results}

\begin{figure}
    \centering
    \includegraphics[width=0.48\textwidth]{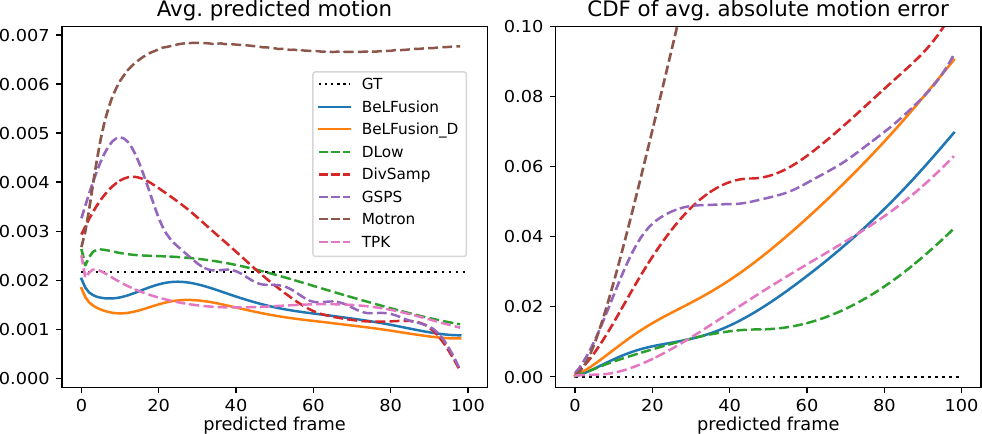}
    \vspace{-0.5cm}
    \caption{Left. Average predicted motion of state-of-the-art methods in H36M. Right. Cumulative distribution function (CDF) of the weighted absolute errors in the left with respect to the ground truth. CMD is the area under this curve.}
    \label{fig:motion_metric_h36m}
    \vspace{-0.4cm}
\end{figure}

\textbf{Comparison with the state of the art.} As shown in \autoref{tab:sota_comparison}, \modelname{} achieves state-of-the-art performance in all accuracy metrics for both datasets. The improvements are especially important in the cross-dataset AMASS configuration, proving its superior robustness against domain shifts. We hypothesize that such good generalization capabilities are due to 1) the exhaustive coverage of behaviors modeled in the disentangled latent space, and 2) the potential of LDMs to model the conditional distribution of future behaviors. In fact, after a single denoising step, our model already achieves state-of-the-art uni- and multimodal ADE and FDE (\modelname{}\_D) in return for less diversity and realism.
When going through all denoising steps (\modelname{}), our method also excels at realism-related metrics like CMD and FID (see the \textit{Implicit diversity} section below for a detailed discussion on the topic).
By contrast, \autoref{fig:motion_metric_h36m} shows that predictions from GSPS and DivSamp consistently accelerate at the beginning, presumably toward divergent poses that promote high diversity values. As a result, they yield high CMD values, especially for H36M.  
The predictions from methods that leverage transformations in the frequency space freeze at the very long-term horizon. 
Motron's high CMD depicts an important jitter in its predictions, missed by all other metrics.
\modelname{}'s low APDE highlights its good ability to adapt to the observed context. This is achieved thanks to 1) the pretrained encoding of the whole observation window, and 2) the behavior coupling to the \textit{target motion}. In contrast, higher APDE values of GSPS and DivSamp are caused by their tendency toward predicting movements more diverse than those present in the dataset. 
Action- (H36M) and dataset-wise (AMASS) results are included in \supp{} Sec. D.1. 

\autoref{fig:qualitative_examples} displays 10 overlaid predictions over time for three actions from H36M (sitting down, eating, and giving directions), and three datasets from AMASS (DanceDB~\cite{dancedb}
, HUMAN4D~\cite{chatzitofis2020human4d}, and GRAB~\cite{taheri2020grab}). The purpose of this visualization is to confirm the observations made by the CMD and APDE metrics. First, the acceleration of GSPS and DivSamp at the beginning of the prediction leads to extreme poses very fast, abruptly transitioning from the observed motion. Second, it shows the capacity of \modelname{} to adapt the diversity predicted to the context. For example, the diversity of motion predicted while eating focuses on the arms, and does not include holistic extreme poses. Interestingly, when just sitting, the predictions include a wider range of full-body movements like laying down, or bending over. A similar context fitting is observed in the AMASS cross-dataset scenario. For instance, \modelname{} correctly identifies that the diversity must target the upper body in the GRAB dataset, or the arms while doing a dance step. Examples \textit{in motion} can be found in \supp{} Sec. E.


\textbf{Ablation study.} Here, we analyze the effect of each of our contributions in the final model quantitatively. This includes the contributions of $\mathcal{L}_{lat}$ and $\mathcal{L}_{rec}$, and the benefits of disentangling behavior from motion in the latent space construction. Results are summarized in \autoref{tab:ablation}. Although training is stable and losses decrease similarly in all cases, solely considering the loss at the coordinate space ($\lossrec$) leads to poor generalization capabilities. This is especially noticeable in the cross-dataset scenario, where models with both latent space constructions are the least accurate among all loss configurations. We observe that the latent loss ($\losslat$) boosts the metrics in both datasets, and can be further enhanced when considered along with the reconstruction loss. 
Overall, the BLS construction benefits all loss configurations in terms of accuracy on both datasets, proving it a very promising strategy to be further explored in HMP.

\begin{figure*}[t!]
    \centering
    \includegraphics[width=\textwidth]{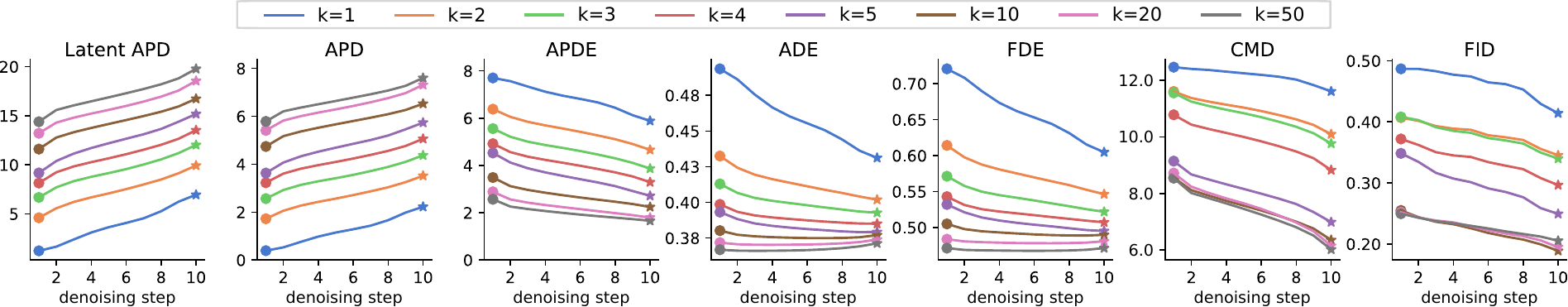}
    \vspace{-0.4cm}
    \caption{Evolution of evaluation metrics (y-axis) along denoising steps (x-axis) at inference time, for different values of $k$. Early stopping can be applied at any time, between the first ($\bullet$) and the last step ($\star$). Accuracy saturates at $k=50$, with gains for all metrics when increasing $k$, especially for diversity (APD). Qualitative metrics (CMD, FID) decrease after each denoising step across all $k$ values.}
    \label{fig:k_analysis}
    \vspace{-0.1cm}
\end{figure*}

\begin{table*}[t!]
    \footnotesize\renewcommand{\arraystretch}{0.9}
    \centering
    \begin{tabular}{ccc@{\hskip 8mm}cccccc@{\hskip 8mm}ccccc}
        \toprule
        & & & \multicolumn{6}{c}{Human3.6M \cite{ionescu2013h36m}} &  \multicolumn{5}{c}{AMASS \cite{mahmood2019amass}} \\
        \toprule
         BLS & $\mathcal{L}_{lat}$ & $\mathcal{L}_{rec}$ & APD & APDE & ADE & FDE & CMD & FID & APD & APDE & ADE & FDE & CMD \\
         \midrule
         
& & \checkmark & \textbf{7.622} & \textbf{1.276} & 0.510 & 0.795 & 5.110 & 2.530 & \textbf{10.788} & 3.032 & 0.697 & 0.881 & \textbf{16.628}\\
\checkmark & & \checkmark & 6.169 & 2.240 & 0.386 & 0.505 & 8.432 & 0.475 & \underline{9.555} & 2.216 & 0.593 & 0.685 & 17.036\\
& \checkmark & & 7.475 & 1.773 & 0.388 & 0.490 & \textbf{4.643} & \textbf{0.177} & 8.688 & 2.079 & 0.528 & 0.572 & 18.429\\
\checkmark & \checkmark & & 6.760 & 1.974 & \underline{0.377} & 0.485 & 6.615 & 0.233 & 8.885 & \underline{2.009} & \underline{0.516} & \underline{0.565} & 17.576 \\
& \checkmark & \checkmark & 7.301 & 2.012 & 0.380 & \underline{0.484} & \underline{4.870} & \underline{0.195} & 8.832 & 2.034 & 0.519 & 0.568 & 17.618 \\
\checkmark & \checkmark & \checkmark & \underline{7.602} & \underline{1.662} & \textbf{0.372} & \textbf{0.474} & 5.988 & 0.209 & 9.376 & \textbf{1.977} & \textbf{0.513} & \textbf{0.560} & \underline{16.995} \\

         \bottomrule
    \end{tabular}
    \vspace{0.1cm}
    \caption{Results from the ablation analysis of \modelname{}. We assess the contribution of the latent ($\mathcal{L}_{lat}$) and reconstruction ($\mathcal{L}_{rec}$) losses, as well as the benefits of applying latent diffusion to a disentangled behavioral latent space (BLS).}
    \label{tab:ablation}
    \vspace{-0.4cm}
\end{table*}


\textbf{Implicit diversity.} As explained in \autoref{subsec:behavioral_ld}, the parameter $k$ regulates the \textit{relaxation} of the training loss (\autoref{eq:final_loss}) on BeLFusion. \autoref{fig:k_analysis} shows how metrics behave when 1) tuning $k$, and 2) moving forward in the reverse diffusion chain (i.e., progressively applying denoising steps). In general, increasing $k$ enhances the samples' diversity, accuracy, and realism. For $k {\leq} 5$, going through the whole chain of denoising steps boosts accuracy. However, for $k {>} 5$, further denoising only boosts diversity- and realism-wise metrics (APD, CMD, FID), and makes the fast single-step inference very accurate. With large enough $k$ values, the LDM learns to cover the conditional space of future behaviors to a great extent and can therefore make a fast and reliable first prediction. The successive denoising steps refine such approximations at expenses of larger inference time. Thus, each denoising step 1) promotes diversity within the latent space, and 2) brings the predicted latent code closer to the true behavioral distribution. Both effects can be observed in the latent APD and FID plots in \autoref{fig:k_analysis}. The latent APD is equivalent to the APD in the latent space of predictions and is computed likewise. 
Note that these effects are not favored by neither the loss choice nor the BLS (see \supp{} Fig. G). Concurrent works have also highlighted the good performance achievable by single-step denoising~\cite{bansal2022colddiff, chen2022diffusiondet}.

\setlength{\tabcolsep}{3pt}
\begin{table}[t!]\renewcommand{\arraystretch}{0.9}
    \footnotesize
    \centering
    \begin{tabular}{l@{\hskip 4mm}cccc}
        \toprule
        & \multicolumn{2}{c}{Human3.6M\cite{ionescu2013h36m}} & \multicolumn{2}{c}{AMASS\cite{mahmood2019amass}} \\
        \midrule
        & Avg. rank & Ranked 1st & Avg. rank & Ranked 1st \\
        \midrule
        GSPS & 2.246 $\pm$ 0.358 & 17.9\% & 2.003 $\pm$ 0.505 & 30.5\% \\
        DivSamp & 2.339 $\pm$ 0.393 & 13.4\% & 2.432 $\pm$ 0.408 & 14.0\% \\
        \modelname{} & \textbf{1.415 $\pm$ 0.217} & \textbf{68.7\%} & \textbf{1.565 $\pm$ 0.332} & \textbf{55.5\%} \\
        \bottomrule
    \end{tabular}
    \vspace{-0.1cm}
    \caption{Qualitative study. \participants{} participants ranked sets of samples from GSPS, DivSamp, and \modelname{} by their realism. Lower average rank ($\pm$ std. dev.) is better.}
    \label{tab:mos_results}\vspace{-0.5cm}
\end{table}
\setlength{\tabcolsep}{6pt}

\textbf{Qualitative assessment.} We performed a qualitative study to assess the realism of \modelname{}'s predictions compared to those of the most accurate methods: DivSamp and GSPS. For each method, we sampled six predictions for 24 randomly sampled observation segments from each dataset (48 in total). We then generated a \textit{gif} that showed both the observed and predicted sequences of the six predictions at the same time. Each participant was asked to order the three sets according to the average realism of the samples. Four questions from either H36M or AMASS were asked to each participant (see \supp{} Sec. F). 
A total of \participants{} people participated in the study. The statistical significance of the results was assessed with the Friedman and Nemenyi tests. Results are shown in \autoref{tab:mos_results}. \modelname{}'s predictions are significantly more realistic than both competitors' in both datasets (p${<}0.01$). GSPS could only be proved significantly more realistic than DivSamp for AMASS (p${<}0.01$). Interestingly, the participant-wise average realism ranks of each method are highly correlated to each method's CMD ($r$=0.730, and $r$=0.601) and APDE ($r$=0.732, and $r$=0.612), for both datasets (H36M, and AMASS, respectively), in terms of Pearson's correlation (p${<}0.001$).

\vspace{-0.3cm}
\section{Conclusion}
\label{sec:conclusions}
\vspace{-0.2cm}
We presented \modelname{}, a latent diffusion model that exploits a behavioral latent space to make more realistic, accurate, and context-adaptive human motion predictions. \modelname{} takes a major step forward in the cross-dataset AMASS configuration. This suggests the necessity of future work to pay attention to domain shifts. These are present in any on-the-wild scenario and therefore on our way toward making highly capable predictive systems.

\textbf{Limitations and future work.} Although sampling with \modelname{} only takes 10 denoising steps, this is still slower than sampling from GANs or VAEs (see \supp{} Sec. D.3.). This may limit its applicability to a real-life scenario. Future work includes exploring our method's capabilities for exploiting a longer observation time-span, and for being auto-regressively applied to predict longer-term sequences.

\textbf{Acknowledgements. }
This work has been partially supported by the Spanish project PID2022-136436NB-I00 and by ICREA under the ICREA Academia programme.

{\small
\bibliographystyle{ieee_fullname}
\bibliography{egbib}
}

\clearpage
\noindent {\Large \textbf{Supplementary Material}\par}
\vspace{0.5cm}
In this supplementary material, we first include additional implementation details to those provided in \autoref{subsec:evaluation_setup} needed to reproduce our work (\autoref{sec:supp_implementation_details}). Then, we complement \autoref{subsec:evaluation_setup} by providing all the information needed to follow the proposed cross-dataset AMASS evaluation protocol (\autoref{sec:supp_dataset_details}). \autoref{subsec:behavioral_ld} is also extended with a 2D visualization of the disentangled behavioral latent space, and several video examples of behavioral transference (\autoref{sec:supp_behavior_transference}). Class- and dataset-wise results from \autoref{subsec:results} are included and discussed (\autoref{sec:supp_exp_results}), as well as a detailed discussion on several video examples comparing \modelname{} against the state of the art (\autoref{sec:supp_visual_examples}). Finally, we provide a thorough description and extended results of the qualitative assessment presented at the end of \autoref{subsec:results} (\autoref{sec:supp_mos}).

\renewcommand*{\thesection}{\Alph{section}}
\renewcommand*{\thefigure}{\Alph{figure}}
\renewcommand*{\thetable}{\Alph{table}}
\setcounter{section}{0}
\setcounter{table}{0}
\setcounter{figure}{0}
\section{Implementation details}
\label{sec:supp_implementation_details}

\def\obsT{B}
\def\predT{T}
\def\xmotion{\mathbf{x}_{m}}
\def\obs{\mathbf{X}}
\def\pred{\mathbf{Y}}

\def\encoder{\mathcal{E}}
\def\decoder{\mathcal{D}}
\def\ldFunction{f_{\Phi}}
\def\latcode{z}
\def\diffused{\latcode_{t}}
\def\diffusedPrev{\latcode_{t-1}}
\def\diffusedStart{\latcode_{0}}

\def\extPred{\pred_{e}}
\def\bvaeEncParams{\theta}
\def\bvaeDecParams{\phi}
\def\bvaeAuxDecParams{\omega}
\def\bvaeXmotionEncParams{\alpha}
\def\bvaeXmotionEnc{g_{\bvaeXmotionEncParams}}
\def\vaeObsEncParams{\lambda}
\def\vaeObsEnc{h_{\vaeObsEncParams}}
\def\bvaeDec{\mathcal{B}_{\bvaeDecParams}}
\def\bvaeEnc{p_{\bvaeEncParams}}
\def\bvaeAuxDec{\mathcal{A}_{\bvaeAuxDecParams}}

\def\lossrec{\mathcal{L}_{rec}}
\def\losslat{\mathcal{L}_{lat}}

To ensure reproducibility, we include in this section all the details regarding \modelname{}'s architecture and training procedure (\autoref{subsec:supp_implementation_ours}). We also cover the details on the implementation of the state-of-the-art models retrained with AMASS (\autoref{subsec:supp_implementation_sota}). 
We follow the terminology used in Fig. 2 and 3 from the main paper.

Note that we only report the hyperparameter values of the best models. For their selection, we conducted grid searches that included learning rate, losses weights, and most relevant network parameters. Data augmentation for all models consisted in randomly rotating from 0 to 360 degrees around the Z axis and mirroring the body skeleton with respect to the XZ- and YZ-planes. The axis and mirroring planes were selected to preserve the floor position and orientation. All models were trained with the ADAM optimizer with AMSGrad~\cite{reddi2019convergence}, with PyTorch 1.9.1 \cite{paszke2019pytorch} and CUDA 11.1 on a single NVIDIA GeForce RTX 3090. The whole \modelname{} training pipleine was trained in 12h for H36M, and 24h for AMASS.

\subsection{\modelname{}}
\label{subsec:supp_implementation_ours}

\begin{figure}[t!]
    \centering
    \includegraphics[width=\linewidth]{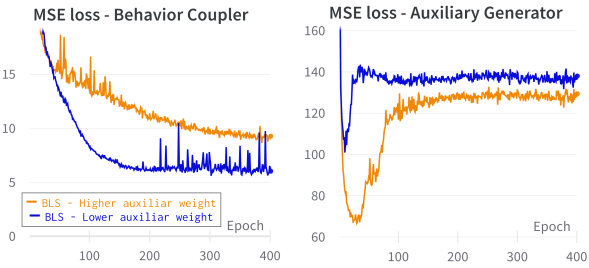}
    \caption{\textbf{Behavioral disentanglement.} Main (left) and adversarial (right) training losses of the behavioral latent space. As expected, when the auxiliary loss weight is higher (orange), the adversarial interplay intensifies.}
    \label{fig:plots_adv_loss}
\end{figure}

\textbf{Behavioral latent space.}
The behavioral VAE consists of four modules. The behavior encoder $\bvaeEnc{}$, which receives the flattened coordinates of all the joints, is composed of a single Gated Recurrent Unit (GRU) cell (hidden state of size 128) followed by a set of a 2D convolutional layers (kernel size of 1, stride of 1, padding of 0) with L2 weight normalization and learned scaling parameters that maps the GRU state to the mean of the latent distribution, and another set to its variance. The behavior coupler $\bvaeDec{}$ consists of a GRU (input shape of 256, hidden state of size 128) followed by a linear layer that maps, at each timestep, its hidden state to the offsets of each joint coordinates with respect to their last observed position. The context encoder $\bvaeXmotionEnc{}$ is an MLP (hidden state of 128) that is fed with the flattened joints coordinates of the target motion $\xmotion$, that includes $C{=}3$ frames. 
Finally, the auxiliary decoder $\bvaeAuxDec{}$ is a clone of $\bvaeDec{}$ with a narrower input shape (128), as only the latent code is fed. Note that the adversarial interplay introduces additional complexity, making convergence more challenging, see \autoref{fig:plots_adv_loss}.
For H36M, the behavioral VAE was trained with learning rates of 0.005 and 0.0005 for $\mathcal{L}_{main}$ and $\mathcal{L}_{aux}$, respectively. For AMASS, they were set to 0.001 and 0.005. All learning rates were decayed with a ratio of 0.9 every 50 epochs. The batch size was set to 64. Each epoch consisted of 5000 and 10000 iterations for H36M and AMASS, respectively. The weight of the $-\mathcal{L}_{aux}$ term in $\mathcal{L}_{main}$ was set to 1.05 for H36M and to 1.00 for AMASS. The KL term was assigned a weight of 0.0001 in both datasets.
Once trained, the behavioral VAE was further fine-tuned for 500 epochs with the behavior encoder $\bvaeEnc{}$ frozen, to enhance the reconstruction capabilities without modifying the disentangled behavioral latent space.
Note that for the ablation study, the non-behavioral latent space was built likewise by disabling the adversarial training framework, and optimizing the model only with the log-likelihood and KL terms of $\mathcal{L}_{main}$ (main paper, Eq. 4), as in a traditional VAE framework.

\textbf{Observation encoding. }The observation encoder $\vaeObsEnc{}$ was pretrained as an autoencoder with an L2 reconstruction loss. It consists of a single-cell GRU layer (hidden state of 64) fed with the flattened joints coordinates. The hidden state of the GRU layer is fed to three MLP layers (output sizes of 300, 200, and 64), and then set as the hidden state of the GRU decoder unit (hidden state of size 64). The sequence is reconstructed by predicting the offsets with respect to the last observed joint coordinates.

\begin{figure}[t!]
    \centering
    \includegraphics[width=0.9\linewidth]{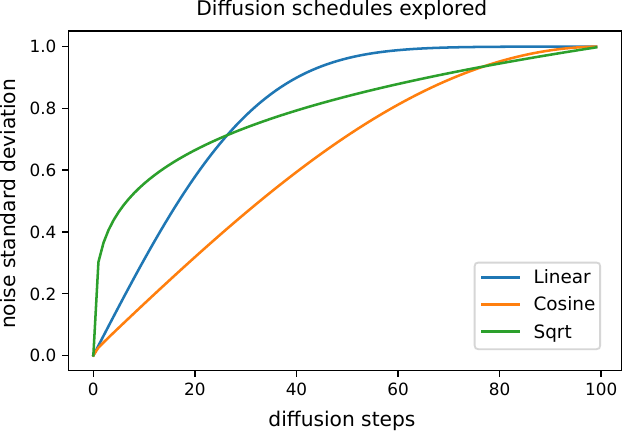}
    \caption{\textbf{Diffusion schedules. } Schedules explored for diffusing the target latent codes.}
    \label{fig:supp_diff_schedules}
\end{figure}

\textbf{Latent diffusion model.} 
\modelname{}'s LDM borrowed its U-Net from \cite{dhariwal2021diffusionbeatsgans}. To leverage it, the target latent codes were reshaped to a rectangular shape (16x8), as prior work proposed~\cite{bautista2022gaudi}. In particular, our U-Net has 2 attention layers (resolutions of 8 and 4), 16 channels per attention head, a FiLM-like conditioning mechanism~\cite{perez2018film}, residual blocks for up and downsampling, and a single residual block. Both the observation and target behavioral encodings were normalized between \text{-1} and 1. 
The LDM was trained with the \textit{sqrt} noise schedule ($s=0.0001$) proposed in \cite{li2022diffusionLM}, which also provided important improvements in our scenario compared to the classic \textit{linear} or \textit{cosine} schedules (see~\autoref{fig:supp_diff_schedules}). With this schedule, the diffusion process is started with a higher noise level, which increases rapidly in the middle of the chain.
The length of the Markov diffusion chain was set to 10, the batch size to 64, the learning rate to 0.0005, and the learning rate decay to a rate of 0.9 every 100 epochs. Each epoch included 10000 samples in both H36M and AMASS training scenarios. Early stopping with a patience of 100 epochs was applied to both, and the epoch where it was triggered was used for the final training with both validation and training sets together. Thus, \modelname{} was trained for 217 epochs in H36M and 1262 for AMASS. For both datasets, the LDM was trained with an exponential moving average (EMA) with a decay of 0.999, triggered every 10 batch iterations, and starting after 1000 initial iterations. The EMA helped reduce the overfitting in the last denoising steps. Predictions were inferred with DDIM sampling~\cite{song2021ddim}. 

\subsection{State-of-the-art models}
\label{subsec:supp_implementation_sota}

The publicly available codes from TPK, DLow, GSPS, and DivSamp were adapted to be trained and evaluated under the AMASS cross-dataset protocol. The best values for their most important hyperparameters were found with grid search. The number of iterations per epoch for all of them was set to 10000.

TPK's loss weights were set to 1000 and 0.1 for the transition and KL losses, respectively. The learning rate was set to 0.001. DLow was trained on top of the TPK model with a learning rate of 0.0001. Its reconstruction and diversity losses weights were set to 2 and 25. For GSPS, the upper- and lower-body joint indices were adapted to the AMASS skeleton configuration. The multimodal ground truth was generated with an upper L2 distance of 0.1, and a lower APD threshold of 0.3. The body angle limits were re-computed with the AMASS statistics. The GSPS learning rate was set to 0.0005, and the weights of the upper- and lower-body diversity losses were set to 5 and 10, respectively. For DivSamp, we used the multimodal ground truth from GSPS, as for H36M they originally borrowed such information from GSPS. For the first training stage (VAE), the learning rate was set to 0.001, and the KL weight to 1. For the second training stage (sampling model), the learning rate was set to 0.0001, the reconstruction loss weight was set to 40, and the diversity loss weight to 20.
For all of them, unspecified parameters were set to the values reported in their original H36M implementations.

\section{AMASS cross-dataset protocol}
\label{sec:supp_dataset_details}

In this section, we give more details to ensure the reproducibility of the cross-dataset AMASS evaluation protocol.

\textbf{Training splits. } The training, validation, and test splits are based on the official AMASS splits from the original publication~\cite{mahmood2019amass}. However, we also include the new datasets added afterward and with available SMPL+H model parameters, up to date. Accordingly, the training set contains the ACCAD, BMLhandball, BMLmovi, BMLrub, CMU, EKUT, EyesJapanDataset, KIT, PosePrior, TCDHands, and TotalCapture datasets, and the validation set contains the HumanEva, HDM05, SFU, and MoSh datasets. The remaining datasets are all part of the test set: DFaust, DanceDB, GRAB, HUMAN4D, SOMA, SSM, and Transitions. AMASS datasets showcase a wide range of behaviors at both intra- and inter-dataset levels. For example, DanceDB, GRAB, and BMLhandball contain sequences of dancing, grabbing objects, and sport actions, respectively. Other datasets like HUMAN4D offer a wide intra-dataset variability of behaviors by themselves. As a result, this evaluation protocol represents a very complete and challenge benchmark for HMP.

\begin{figure}[t!]
    \centering
    \includegraphics[width=0.9\linewidth]{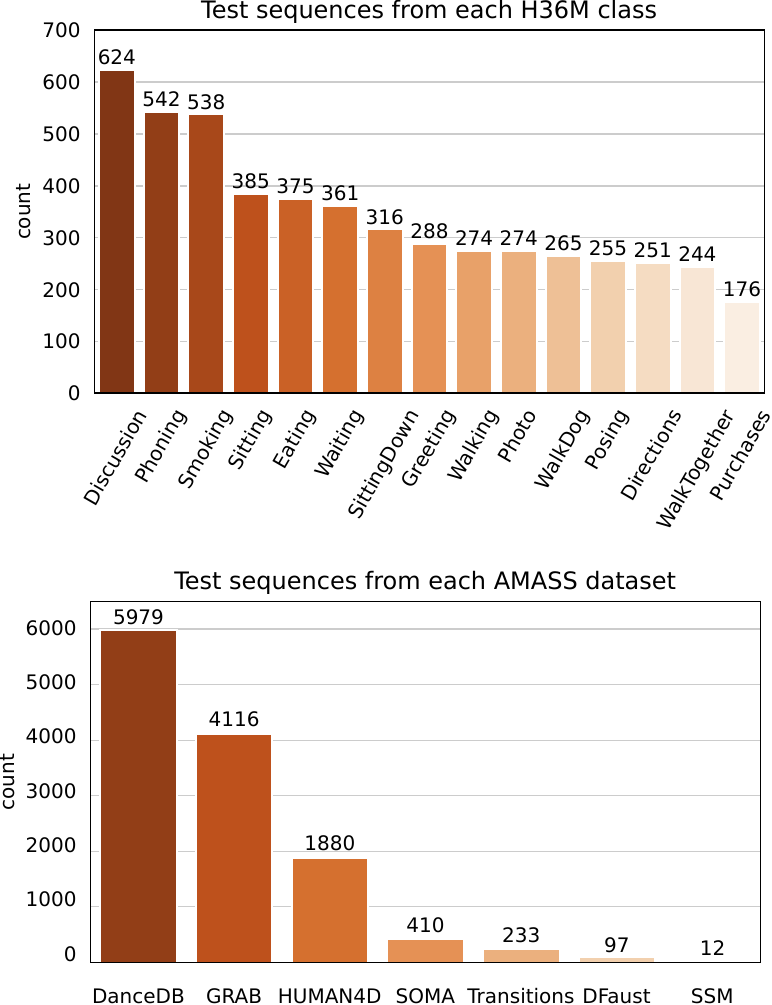}
    \caption{\textbf{Test set sequences. }We show the number of test sequences evaluated for each class/dataset in H36M/AMASS.}
    \vspace{-0.3cm}
    \label{fig:supp_datasets_distribution}
\end{figure}

\textbf{Test sequences. }For each dataset clip (previously downsampled to 60Hz), we selected all sequences starting from frame 180 (3s), with a stride of 120 (2s). This was done to ensure that for any segment to predict (prediction window), up to 3s of preceding motion was available. As a result, future work will be able to explore models exploiting longer observation windows while still using the same prediction windows and, therefore, be compared to our results. A total of 12728 segments were selected, around 2.5 times the amount of H36M test sequences. Note that those clips with no framerate available in AMASS metadata were ignored. \autoref{fig:supp_datasets_distribution} shows the number of segments extracted from each test dataset. 94.1\% of all test samples belong to either DanceDB, GRAB, or HUMAN4D. Most SSM clips had to be discarded due to lengths shorter than 300 frames (5s). The list of sequence indices is made available along the project code for easing reproducibility.

\textbf{Multimodal ground truth. }The L2 distance threshold used for the generation of the multimodal ground truth was set to 0.4 so that the average number of resulting multimodal ground truths for each sequence was similar to that of H36M with a threshold of 0.5~\cite{yuan2020dlow}. %

\section{Behavioral latent space}
\label{sec:supp_behavior_transference}

In this section, we present 1) a t-SNE plot for visualizing the behavioral latent space of the H36M test segments, and 2) visual examples of transferring behavior to ongoing motions.

\begin{figure}
    \centering
    \includegraphics[width=0.48\textwidth]{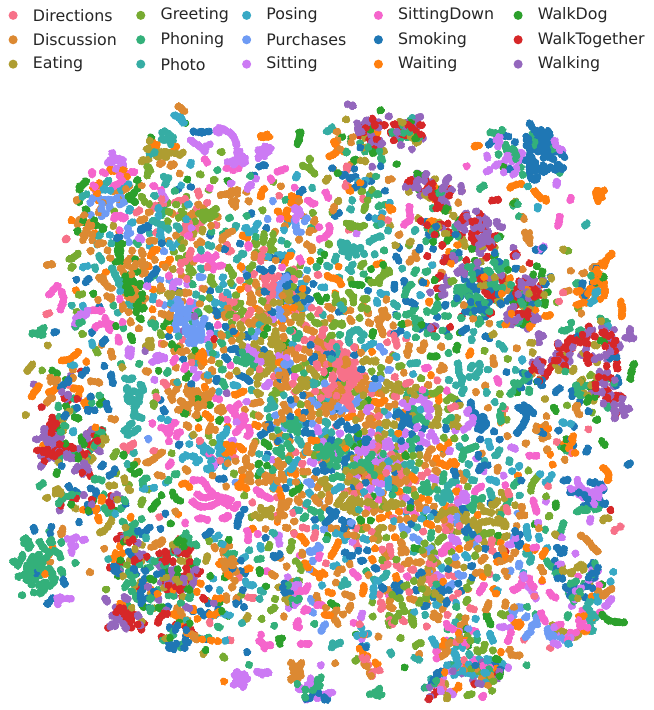}
    \caption{\textbf{Behavioral latent space. }2D projection of the behavioral encodings of all H36M test sequences generated with t-SNE.}
    \label{fig:supp_tsne_behaviors}
\end{figure}

\textbf{2D projection. }\autoref{fig:supp_tsne_behaviors} shows a 2-dimensional t-SNE projection of all behavioral encodings of the H36M test sequences~\cite{van2008tsne}. Note that, despite its class label, a sequence may show actions of another class. For example, \textit{Waiting} sequences include sub-sequences where the person walks or sits down. Interestingly, we can observe that most walking-related sequences (\textit{WalkDog}, \textit{WalkTogether}, \textit{Walking}) are clustered together in the top-right and bottom-left corners. Such entanglement within those clusters suggests that the task of choosing the way to keep walking might be relegated to the behavior coupler, which has information on how the action is being performed. Farther in those corners, we can also find very isolated clusters of \textit{Phoning} and \textit{Smoking}, whose proximity to the walking behaviors suggests that such sequences may involve a subject making a call or smoking while walking. However, without fine-grained annotations at the sequence level, we cannot come to any strong conclusion.

\textbf{Transference of behaviors. }We include several videos\footnote{Videos referenced in the \supp{} can be found in: \small\url{https://barquerogerman.github.io/BeLFusion/.}} showing the capabilities of the behavior coupler to transfer a behavioral latent code to any ongoing motion. The motion tagged as \textit{behavior} shows the target behavior to be encoded and transferred. All the other columns show the ongoing motions where the behavior will be transferred to. They are shown with blue and orange skeletons. Once the behavior is transferred, the color of the skeletons switches to green and pink. In `H1' (H36M), the walking action or behavior is transferred to the target ongoing motions. For ongoing motions where the person is standing, they start walking towards the direction they are facing (\#1, \#2, \#4, \#5). Such transition is smooth and coherent with the observation. For example, the person making a phone call in \#7 keeps the arm next to the ear while starting to walk. When sitting or bending down, the movement of the legs is either very little (\#3 and \#6), or very limited (\#8). `H2' and `H3' show the transference of subtle and long-range behaviors, respectively. For AMASS, such behavioral encoding faces a huge domain drift. However, we still observe good results at this task. For example, `A1' shows how a \textit{stretching} movement is successfully transferred to very distinct ongoing motions by generating smooth and realistic transitions. Similarly, `A2' and `A3' are examples of transferring subtle and aggressive behaviors, respectively. Even though the dancing behavior in `A3' was not seen at training time, it is transferred and adapted to the ongoing motion fairly realistically.

\setlength{\tabcolsep}{1.4pt}
\begin{table*}[t!]\renewcommand{\arraystretch}{0.9}
\footnotesize		
\centering

\begin{minipage}{0.495\textwidth}
\begin{tabular}{lcccccccc@{\hskip 2mm}}
\toprule
Classes & APD & APDE  & ADE  & FDE  & MMADE  & MMFDE & CMD & FID\\

\midrule

Directions \\
\midrule
TPK & 6.510 & \underline{2.039} & 0.447 & 0.482 & 0.523 & 0.544 & \underline{7.455} & \underline{1.768}\\
DLow & 11.874 & 3.359 & 0.415 & 0.465 & 0.499 & \underline{0.514} & \textbf{2.011} & 4.633 \\
GSPS & \underline{15.398} & 6.877 & 0.407 & 0.477 & \underline{0.492} & 0.522 & 10.469 & 4.827 \\
DivSamp & \textbf{15.663} & 7.142 & \underline{0.389} & \underline{0.463} & 0.502 & 0.523 & 10.539 & 5.489 \\
BeLFusion & 7.090 & \textbf{1.709} & \textbf{0.378} & \textbf{0.422} & \textbf{0.484} & \textbf{0.494} & 10.110 & \textbf{1.150} \\
\midrule

Discussion \\
\midrule
TPK & 6.966 & \underline{2.572} & 0.511 & 0.581 & 0.570 & 0.600 & 7.554 & \underline{1.090} \\
DLow & 11.872 & 2.659 & 0.472 & 0.536 & 0.533 & \underline{0.549} & \textbf{2.695} & 1.300 \\
GSPS & \underline{14.199} & 4.992 & 0.448 & 0.541 & \underline{0.526} & 0.563 & 8,470 & 1.870 \\
DivSamp & \textbf{15.310} & 5.905 & \underline{0.432} & \underline{0.526} & 0.534 & 0.557 & 8.975 & 1.522 \\
BeLFusion & 9.172 & \textbf{1.425} & \textbf{0.420} & \textbf{0.507} & \textbf{0.512} & \textbf{0.530} & \underline{7.521} & \textbf{1.055} \\
\midrule

Eating \\
\midrule
TPK & 6.412 & \textbf{1.066} & 0.388 & 0.473 & 0.452 & 0.472 & \underline{5.306} & \underline{4.345} \\
DLow & 11.603 & 4.829 & 0.358 & 0.433 & 0.439 & 0.452 & \textbf{3.214} & 10.300 \\
GSPS & \underline{15.570} & 8.793 & 0.334 & \underline{0.419} & \underline{0.424} & 0.448 & 12.360 & 11.322 \\
DivSamp & \textbf{15.681} & 8.904 & \underline{0.321} & \underline{0.419} & 0.428 & \underline{0.445} & 13.863 & 10.270 \\
BeLFusion & 5.954 & \underline{1.297} & \textbf{0.310} & \textbf{0.381} & \textbf{0.418} & \textbf{0.420} & 5.808 & \textbf{1.439} \\
\midrule

Greeting \\
\midrule
TPK & 6.779 & \underline{2.545} & 0.555 & 0.615 & 0.571 & 0.598 & 12.313 & \textbf{2.148} \\
DLow & 11.897 & 3.112 & 0.530 & 0.590 & 0.542 & 0.564 & \textbf{5.994} & 3.724 \\
GSPS & \underline{14.974} & 5.950 & 0.502 & 0.592 & \underline{0.532} & 0.577 & 10.654 & 5.488 \\
DivSamp & \textbf{15.447} & 6.373 & \underline{0.489} & \underline{0.579} & 0.535 & \underline{0.562} & \underline{9.044} & 4.848 \\
BeLFusion & 8.482 & \textbf{1.690} & \textbf{0.482} & \textbf{0.544} & \textbf{0.524} & \textbf{0.540} & 12.740 & \underline{2.201} \\
\midrule

Phoning \\
\midrule
TPK & 6.410 & \textbf{1.400} & 0.377 & 0.475 & 0.468 & 0.507 & \textbf{3.057} & \underline{1.882} \\
DLow & 11.542 & 4.605 & 0.343 & 0.444 & 0.451 & 0.487 & 4.886 & 4.847 \\
GSPS & \underline{15.050} & 8.120 & 0.311 & 0.413 & \underline{0.436} & 0.476 & 12.292 & 6.458 \\
DivSamp & \textbf{15.751} & 8.813 & \underline{0.296} & \underline{0.400} & 0.437 & \underline{0.471} & 14.295 & 5.149 \\
BeLFusion & 6.649 & \underline{1.477} & \textbf{0.283} & \textbf{0.375} & \textbf{0.426} & \textbf{0.445} & \underline{3.388} & \textbf{0.836} \\
\midrule

Photo \\
\midrule
TPK & 6.894 & \underline{1.884} & 0.541 & 0.689 & 0.548 & 0.633 & \textbf{3.928} & \underline{3.231} \\
DLow & 11.931 & 4.180 & 0.507 & \underline{0.655} & 0.516 & \underline{0.596} & \underline{4.013} & 3.305 \\
GSPS & \underline{14.310} & 6.482 & 0.485 & 0.663 & \underline{0.502} & 0.606 & 10.855 & 3.851 \\
DivSamp & \textbf{15.330} & 7.428 & \underline{0.474} & 0.665 & 0.506 & 0.607 & 11.427 & 4.571 \\
BeLFusion & 8.446 & \textbf{1.726} & \textbf{0.434} & \textbf{0.601} & \textbf{0.462} & \textbf{0.546} & 4.491 & \textbf{2.526} \\
\midrule

Posing \\
\midrule
TPK & 6.520 & \underline{2.310} & 0.466 & 0.538 & 0.542 & 0.565 & 4.740 & \textbf{1.279} \\
DLow & 11.875 & 3.116 & 0.442 & 0.521 & 0.510 & \textbf{0.525} & \textbf{3.421} & 2.521 \\
GSPS & \underline{15.149} & 6.399 & 0.415 & 0.527 & \textbf{0.498} & 0.543 & 10.720 & 4.967 \\
DivSamp & \textbf{15.429} & 6.676 & \textbf{0.395} & \textbf{0.499} & 0.510 & 0.541 & 11.201 & 4.143 \\
BeLFusion & 8.438 & \textbf{1.241} & \underline{0.406} & \underline{0.510} & \textbf{0.498} & \underline{0.531} & \underline{4.729} & \underline{1.463} \\
\midrule

Purchases \\
\midrule
TPK & 7.450 & \underline{2.161} & 0.505 & 0.522 & 0.535 & 0.538 & 10.298 & 7.194 \\
DLow & 11.947 & 2.629 & 0.430 & 0.422 & \textbf{0.493} & \underline{0.477} & \textbf{5.090} & 6.871 \\
GSPS & \underline{13.969} & 4.552 & 0.414 & 0.429 & 0.497 & 0.497 & 7.380 & 6.521 \\
DivSamp & \textbf{14.967} & 5.517 & \textbf{0.388} & \textbf{0.404} & 0.502 & 0.478 & \underline{6.950} & \textbf{3.758} \\
BeLFusion & 10.272 & \textbf{1.738} & \underline{0.410} & \underline{0.409} & \underline{0.494} & \textbf{0.472} & 8.800 & \underline{5.696} \\
\bottomrule
\end{tabular}

\end{minipage} \hfill
\begin{minipage}{0.495\textwidth}
\vspace{-2.15cm}
\begin{tabular}{lcccccccc@{\hskip 2mm}}
\toprule
Classes & APD & APDE  & ADE  & FDE  & MMADE  & MMFDE & CMD & FID\\

\midrule

Sitting \\
\midrule
TPK & 6.417 & \textbf{1.167} & 0.400 & 0.547 & 0.461 & 0.548 & \textbf{1.542} & \textbf{1.619} \\
DLow & 11.425 & 4.972 & 0.364 & 0.513 & 0.440 & 0.523 & 7.490 & 3.290 \\
GSPS & \underline{14.966} & 8.494 & 0.323 & \underline{0.454} & \underline{0.411} & \underline{0.484} & 14.377 & 5.717 \\
DivSamp & \textbf{15.614} & 9.146 & \underline{0.317} & 0.465 & 0.417 & 0.490 & 16.828 & 3.485 \\
BeLFusion & 6.495 & \underline{1.233} & \textbf{0.306} & \textbf{0.446} & \textbf{0.400} & \textbf{0.461} & \underline{1.957} & \underline{1.836} \\
\midrule

\multicolumn{2}{l}{SittingDown} \\
\midrule
TPK & 7.393 & \textbf{1.864} & 0.496 & 0.678 & 0.531 & 0.666 & \textbf{2.889} & \textbf{1.987} \\
DLow & 12.044 & 4.576 & 0.451 & 0.605 & 0.495 & 0.606 & 5.651 & 2.759 \\
GSPS & \underline{13.725} & 6.520 & \textbf{0.406} & \textbf{0.561} & \textbf{0.461} & \textbf{0.565} & 9.301 & 3.694 \\
DivSamp & \textbf{14.899} & 7.240 & 0.413 & \underline{0.579} & 0.478 & \underline{0.586} & 11.929 & 3.471 \\
BeLFusion & 9.026 & \underline{2.236} & \underline{0.413} & 0.585 & \underline{0.468} & 0.587 & \underline{2.997} & \underline{2.642} \\
\midrule

Smoking \\
\midrule
TPK & 6.522 & \underline{1.807} & 0.422 & 0.529 & 0.509 & 0.560 & \textbf{3.148} & \underline{1.652} \\
DLow & 11.549 & 4.058 & 0.400 & 0.515 & 0.490 & 0.537 & 5.123 & 3.535 \\
GSPS & \underline{14.822} & 7.332 & 0.366 & \underline{0.485} & \underline{0.472} & 0.530 & 11.478 & 4.622 \\
DivSamp & \textbf{15.688} & 8.153 & \underline{0.353} & 0.486 & 0.475 & \underline{0.523} & 14.041 & 4.258 \\
BeLFusion & 6.780 & \textbf{1.372} & \textbf{0.341} & \textbf{0.467} & \textbf{0.467} & \textbf{0.512} & \underline{3.849} & \textbf{0.847} \\
\midrule

Waiting \\
\midrule
TPK & 6.631 & \underline{2.080} & 0.480 & 0.584 & 0.526 & 0.568 & \underline{4.143} & \underline{1.022} \\
DLow & 11.680 & 3.398 & 0.441 & 0.541 & 0.497 & 0.534 & \textbf{3.866} & 1.758 \\
GSPS & \underline{15.000} & 6.702 & 0.400 & \underline{0.514} & \underline{0.475} & \underline{0.529} & 10.686 & 3.277 \\
DivSamp & \textbf{15.455} & 7.156 & \textbf{0.387} & 0.517 & 0.486 & 0.535 & 11.611 & 3.108 \\
BeLFusion & 7.747 & \textbf{1.542} & \underline{0.390} & \textbf{0.507} & \textbf{0.471} & \textbf{0.511} & 4.186 & \textbf{0.981} \\
\midrule

WalkDog \\
\midrule
TPK & 7.384 & \underline{2.481} & 0.560 & 0.694 & 0.592 & 0.665 & 13.157 & 3.395 \\
DLow & 11.882 & 2.732 & 0.490 & 0.566 & 0.539 & \underline{0.570} & \underline{8.495} & \underline{3.019} \\
GSPS & \underline{13.746} & 4.569 & 0.459 & 0.564 & \underline{0.530} & 0.587 & 8.869 & 2.647 \\
DivSamp & \textbf{15.616} & 6.212 & \underline{0.439} & \underline{0.555} & 0.532 & 0.577 & \textbf{8.177} & \textbf{1.979} \\
BeLFusion & 9.335 & \textbf{1.893} & \textbf{0.432} & \textbf{0.530} & \textbf{0.527} & \textbf{0.569} & 11.908 & 3.193 \\
\midrule

\multicolumn{2}{l}{WalkTogether} \\
\midrule
TPK & 6.718 & \textbf{1.791} & 0.443 & 0.548 & 0.535 & 0.573 & 10.814 & \underline{14.715} \\
DLow & 11.951 & 3.922 & 0.395 & 0.495 & 0.503 & 0.530 & \textbf{5.234} & 20.315 \\
GSPS & \underline{15.030} & 6.994 & \underline{0.316} & \underline{0.440} & \textbf{0.473} & \underline{0.516} & 10.265 & 22.212 \\
DivSamp & \textbf{16.095} & 8.060 & 0.321 & 0.458 & 0.486 & 0.525 & 10.584 & 19.643 \\
BeLFusion & 6.378 & \underline{2.092} & \textbf{0.296} & \textbf{0.393} & \underline{0.484} & \textbf{0.495} & \underline{5.613} & \textbf{4.348} \\
\midrule

Walking \\
\midrule
TPK & 6.708 & \textbf{1.875} & 0.455 & 0.533 & 0.538 & 0.558 & 14.279 & \underline{16.210} \\
DLow & 11.904 & 3.507 & 0.428 & 0.518 & 0.516 & \underline{0.539} & \textbf{8.400} & 20.670 \\
GSPS & \underline{14.797} & 6.399 & \textbf{0.351} & \textbf{0.469} & \textbf{0.490} & \textbf{0.528} & 10.352 & 19.394 \\
DivSamp & \textbf{15.964} & 7.566 & 0.373 & 0.535 & \underline{0.508} & 0.547 & 10.024 & 17.166 \\
BeLFusion & 5.116 & \underline{3.345} & \underline{0.367} & \underline{0.471} & 0.530 & 0.546 & \underline{8.588} & \textbf{3.784} \\

 \bottomrule
    \end{tabular}
\end{minipage}
    \caption{Comparison of \modelname{} with state-of-the-art methods on H36M. Bold and underlined results correspond to the best and second-
best results, respectively. Lower is better for all metrics except APD.}
    \label{tab:supp_sota_comparison_h36m}
\end{table*}
\setlength{\tabcolsep}{6pt}

\setlength{\tabcolsep}{2pt}
\begin{table}[t!]\renewcommand{\arraystretch}{0.9}
\footnotesize		
\centering
\begin{tabular}{lccccccc@{\hskip 2mm}}
\toprule
Datasets & APD & APDE  & ADE  & FDE  & MMADE  & MMFDE & CMD\\
\midrule
DFaust \\
\midrule
TPK & 8.998 & \textbf{2.435} & 0.591 & 0.555 & 0.637 & 0.601 & 8.263\\
DLow & 12.805 & 2.755 & 0.521 & 0.505 & 0.565 & \underline{0.539} & \textbf{3.640}\\
GSPS & \underline{12.870} & 3.218 & 0.504 & 0.508 & \underline{0.564} & 0.556 & \underline{8.150}\\
DivSamp & \textbf{25.016} & 14.691 & \underline{0.479} & \underline{0.495} & 0.569 & 0.569 & 57.256\\
BeLFusion & 9.285 & \underline{2.456} & \textbf{0.441} & \textbf{0.424} & \textbf{0.514} & \textbf{0.498} & 14.174\\
\midrule
DanceDB \\
\midrule
TPK & 9.665 & \underline{2.812} & 0.810 & 0.798 & 0.815 & 0.796 & \underline{25.232}\\
DLow & \underline{13.703} & 3.307 & 0.763 & \underline{0.760} & 0.769 & \underline{0.756} & \textbf{18.800}\\
GSPS & 11.792 & 3.121 & \underline{0.747} & 0.764 & \underline{0.758} & 0.765 & 27.113\\
DivSamp & \textbf{23.984} & 13.008 & 0.757 & 0.815 & 0.777 & 0.818 & 31.244\\
BeLFusion & 10.619 & \textbf{2.780} & \textbf{0.690} & \textbf{0.713} & \textbf{0.709} & \textbf{0.717} & 28.874\\
\midrule
GRAB \\
\midrule
TPK & 8.590 & \underline{1.555} & 0.415 & 0.457 & 0.463 & 0.469 & \underline{9.646}\\
DLow & 12.376 & 5.180 & 0.338 & 0.383 & 0.407 & \underline{0.411} & 15.502\\
GSPS & \underline{13.515} & 6.331 & 0.300 & \underline{0.381} & \underline{0.404} & 0.435 & 11.642\\
DivSamp & \textbf{25.882} & 18.686 & \underline{0.287} & 0.394 & 0.407 & 0.447 & 76.817\\
BeLFusion & 7.421 & \textbf{1.111} & \textbf{0.260} & \textbf{0.323} & \textbf{0.375} & \textbf{0.388} & \textbf{1.321}\\
\midrule
HUMAN4D \\
\midrule
TPK & 9.451 & \underline{2.618} & 0.657 & 0.732 & 0.662 & 0.705 & 6.305\\
DLow & \underline{13.083} & 4.571 & 0.562 & 0.629 & 0.583 & \underline{0.612} & \textbf{2.888}\\
GSPS & 12.449 & 4.764 & \underline{0.514} & \underline{0.609} & \underline{0.563} & 0.617 & \underline{4.099}\\
DivSamp & \textbf{24.665} & 16.149 & 0.519 & 0.632 & 0.581 & 0.641 & 57.120\\
BeLFusion & 9.262 & \textbf{2.020} & \textbf{0.471} & \textbf{0.568} & \textbf{0.526} & \textbf{0.576} & 10.909\\
\midrule
SOMA \\
\midrule
TPK & 9.823 & \underline{3.166} & 0.806 & 0.835 & 0.798 & 0.817 & \underline{20.689}\\
DLow & \underline{13.761} & 3.402 & 0.726 & \underline{0.746} & 0.722 & \underline{0.737} & \textbf{15.123}\\
GSPS & 11.867 & 3.665 & \underline{0.715} & 0.779 & \underline{0.710} & 0.765 & 22.222\\
DivSamp & \textbf{24.131} & 13.296 & 0.724 & 0.802 & 0.728 & 0.795 & 35.350\\
BeLFusion & 10.765 & \textbf{3.106} & \textbf{0.647} & \textbf{0.691} & \textbf{0.655} & \textbf{0.685} & 23.727\\
\midrule
SSM \\
\midrule
TPK & 9.459 & \underline{2.741} & 0.595 & 0.486 & 0.662 & 0.615 & 13.479\\
DLow & \underline{13.029} & 3.290 & 0.498 & \underline{0.379} & 0.559 & \textbf{0.466} & \textbf{8.491}\\
GSPS & 12.973 & 3.467 & 0.490 & 0.412 & \underline{0.556} & 0.504 & \underline{12.369}\\
DivSamp & \textbf{24.993} & 14.164 & \underline{0.474} & 0.416 & 0.580 & 0.568 & 56.610\\
BeLFusion & 9.576 & \textbf{1.916} & \textbf{0.433} & \textbf{0.356} & \textbf{0.502} & \underline{0.470} & 19.351\\
\midrule
Transitions \\
\midrule
TPK & 9.525 & \underline{2.217} & 0.696 & 0.672 & 0.706 & 0.658 & \underline{26.234}\\
DLow & \underline{13.308} & 2.461 & \underline{0.599} & \textbf{0.538} & \underline{0.615} & \textbf{0.550} & \textbf{21.308}\\
GSPS & 12.169 & 2.470 & 0.636 & 0.642 & 0.655 & 0.648 & 27.634\\
DivSamp & \textbf{24.612} & 14.092 & 0.648 & 0.724 & 0.687 & 0.725 & 33.953\\
BeLFusion & 10.499 & \textbf{2.085} & \textbf{0.577} & \underline{0.578} & \textbf{0.611} & \underline{0.596} & 27.361\\

 \bottomrule
    \end{tabular}
    \caption{Comparison of \modelname{} with state-of-the-art methods on AMASS. Bold and underlined results correspond to the best and second-
best results, respectively. Lower is better for all metrics except APD.}
    \label{tab:supp_sota_comparison_amass}
\end{table}
\setlength{\tabcolsep}{6pt}

\section{Further experimental results}
\label{sec:supp_exp_results}

In this section, we present a class- and dataset-wise comparison to the state of the art for H36M and AMASS, respectively (\autoref{subsec:supp_wise_results}). We also include the distributions of predicted displacement for each class/dataset, which are used for the CMD calculation. Then, we present an extended analysis of the effect of $k$, which controls the loss \textit{relaxation} level (\autoref{subsec:supp_extended_k}).
Finally, we compare the inference time of \modelname{} to the state of the art (\autoref{subsec:supp_implementation_times}).

\subsection{Class- and dataset-wise results}
\label{subsec:supp_wise_results}

\begin{figure*}[t!]
    \centering
    \includegraphics[width=\textwidth]{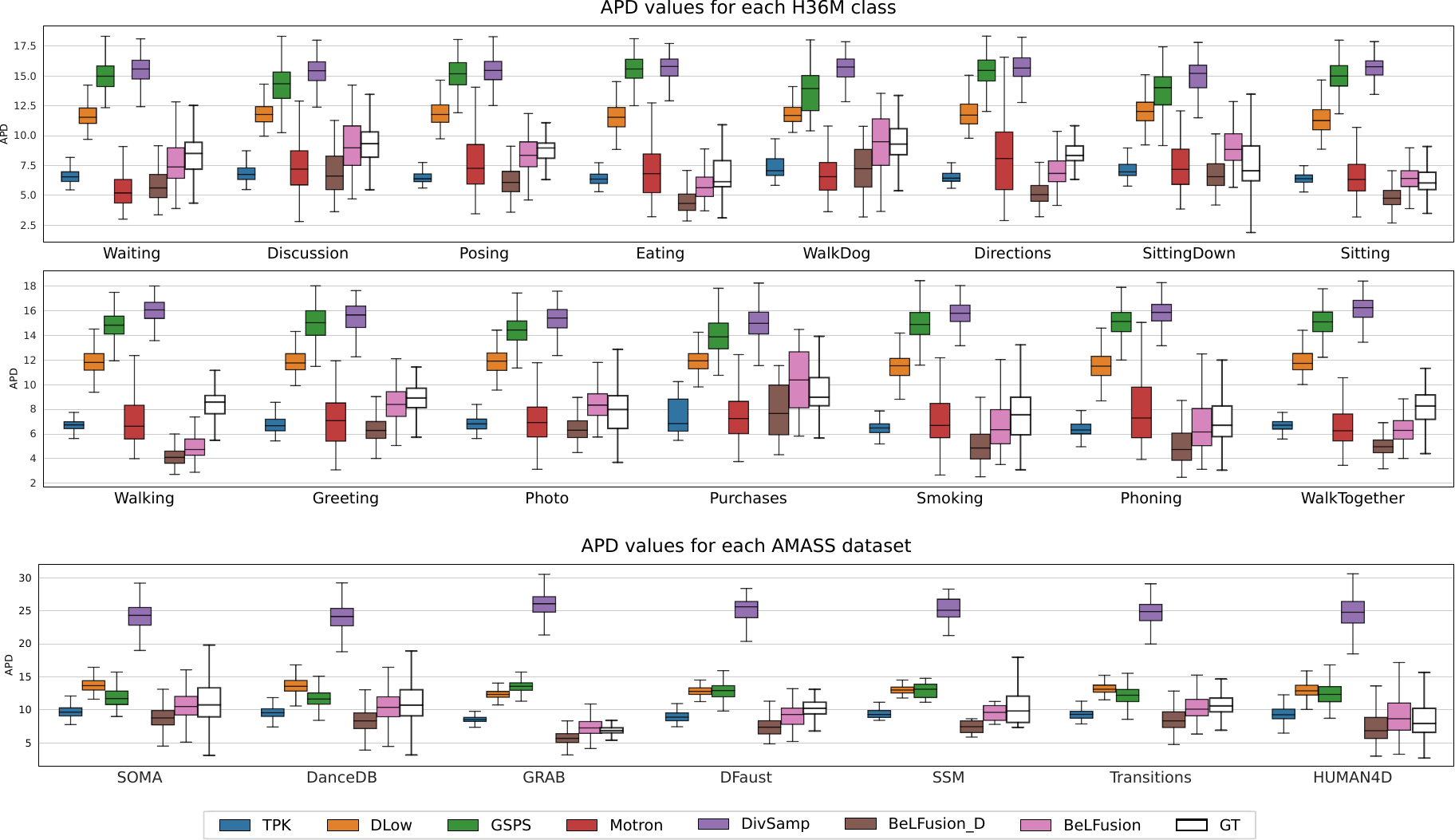}
    \caption{\textbf{Class- and dataset-wise APD. }GT corresponds to the APD of the multimodal ground truth. \modelname{} is the only method that adjusts the diversity of its predictions to model the intrinsic diversity of each class and dataset. As a result, the APD distributions between \modelname{} and GT are very similar. The proposed APDE metric quantifies such similarity (the lower, the more similar).}
    \label{fig:supp_apde}
\end{figure*}

\autoref{tab:supp_sota_comparison_h36m} shows that \modelname{} achieves state-of-the-art results in most metrics in all H36M classes. We stress that our model is especially good at predicting the future in contexts where the observation strongly determines the following action. For example, when the person is \textit{Smoking}, or \textit{Phoning}, a model should predict a coherent future that also involves holding a cigar, or a phone. \modelname{} succeeds at it, showing improvements of 9.1\%, 6.3\%, and 3.7\% for FDE with respect to other methods for \textit{Eating}, \textit{Phoning}, and \textit{Smoking}, respectively. Our model also excels in classes where the determinacy of each part of the body needs to be assessed. For example, for \textit{Directions}, and \textit{Photo}, which often involve a static lower-body, and diverse upper-body movements, \modelname{} improves FDE by an 8.9\%, and an 8.0\%, respectively. We also highlight the adaptive APD that our model shows, in contrast to the constant variety of motions predicted by the state-of-the-art methods. Such effect is better observed in \autoref{fig:supp_apde}, where \modelname{} is the method that best replicates the intrinsic multimodal diversity of each class (i.e., APD of the multimodal ground truth, see Sec. 4.2).
The variety of motions present in each AMASS dataset impedes such a detailed analysis. However, we also observe that the improvements with respect to the other methods are consistent across datasets (\autoref{tab:supp_sota_comparison_amass}). The only dataset where \modelname{} is beaten in an accuracy metric (FDE) is Transitions, where the sequences consist of transitions among different actions, without any behavioral cue that allows the model to anticipate it. We also observe that our model yields a higher variability of APD across datasets that adapts to the sequence context, clearly depicted in \autoref{fig:supp_apde} as well.

\begin{figure*}
    \centering
    \includegraphics[width=\textwidth]{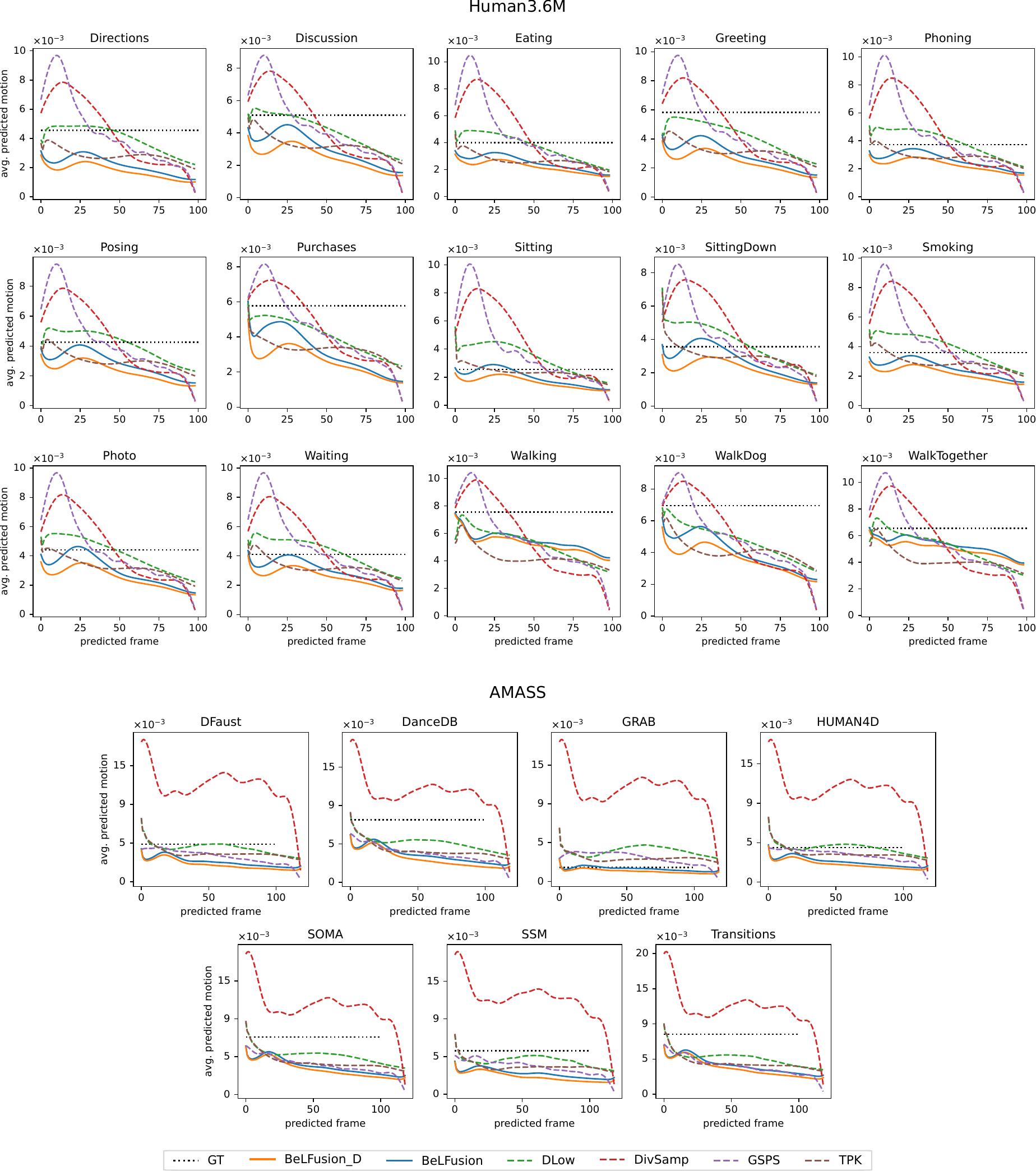}
    \caption{\textbf{Predicted motion analysis. } For each timestep in the future (predicted frame), the plots above show the displacement predicted averaged across all test sequences. For H36M, GSPS and DivSamp predictions accelerate in the beginning, leading to unrealistic transitions. For AMASS, DivSamp shows a similar behavior, and DLow beats all methods except in GRAB, where \modelname{} matches very well the average dataset motion.}
    \label{fig:supp_supp_cmd}
\end{figure*}

Regarding the CMD, Tab.~\ref{tab:supp_sota_comparison_h36m} and \ref{tab:supp_sota_comparison_amass} show how methods that promote highly diverse predictions are biased toward forecasting faster movements than the ones present in the dataset. \autoref{fig:supp_supp_cmd} shows a clearer picture of this bias by plotting the average predicted displacement at all predicted frames. We observe how in all H36M classes, GSPS and DivSamp accelerate very early and eventually stop by the end of the prediction. We argue that such early divergent motion favors high diversity values, at expense of realistic transitions from the ongoing to the predicted motion. By contrast, \modelname{} produces movements that resemble those present in the dataset. While DivSamp follows a similar trend in AMASS than in H36M, GSPS does not. Although DLow is far from state-of-the-art accuracy, it achieves the best performance with regard to this metric in both datasets. Interestingly, \modelname{} slightly decelerates at the first frames and then achieves the motion closest to that of the dataset shortly after. We hypothesize that this effect is an artifact of the behavioral coupling step, where the ongoing motion smoothly transitions to the predicted behavior.

\subsection{Ablation study: implicit diversity} 
\label{subsec:supp_extended_k}

As described in Sec. 3.3 and 4.3 of the main paper, by relaxing the loss regularization (i.e., increasing the number of predictions sampled at each training iteration, $k$), we can increase the diversity of \modelname{}'s predictions. We argue that by backpropagating each loss only on its best prediction out of $k$ (Eq. 6): 1) wrong predictions are not penalized, and 2) correct predictions of less frequent behaviors are rewarded. In the disentangled BLS, distinct behaviors are spread out (\autoref{fig:supp_tsne_behaviors}). Thus, high $k$'s implicitly encourage models denoising $\mathcal{N}(0, 1)$ into a distribution with multiple behavioral modes (i.e., into behaviorally diverse futures).

\begin{figure*}
    \centering
    \includegraphics[width=\textwidth]{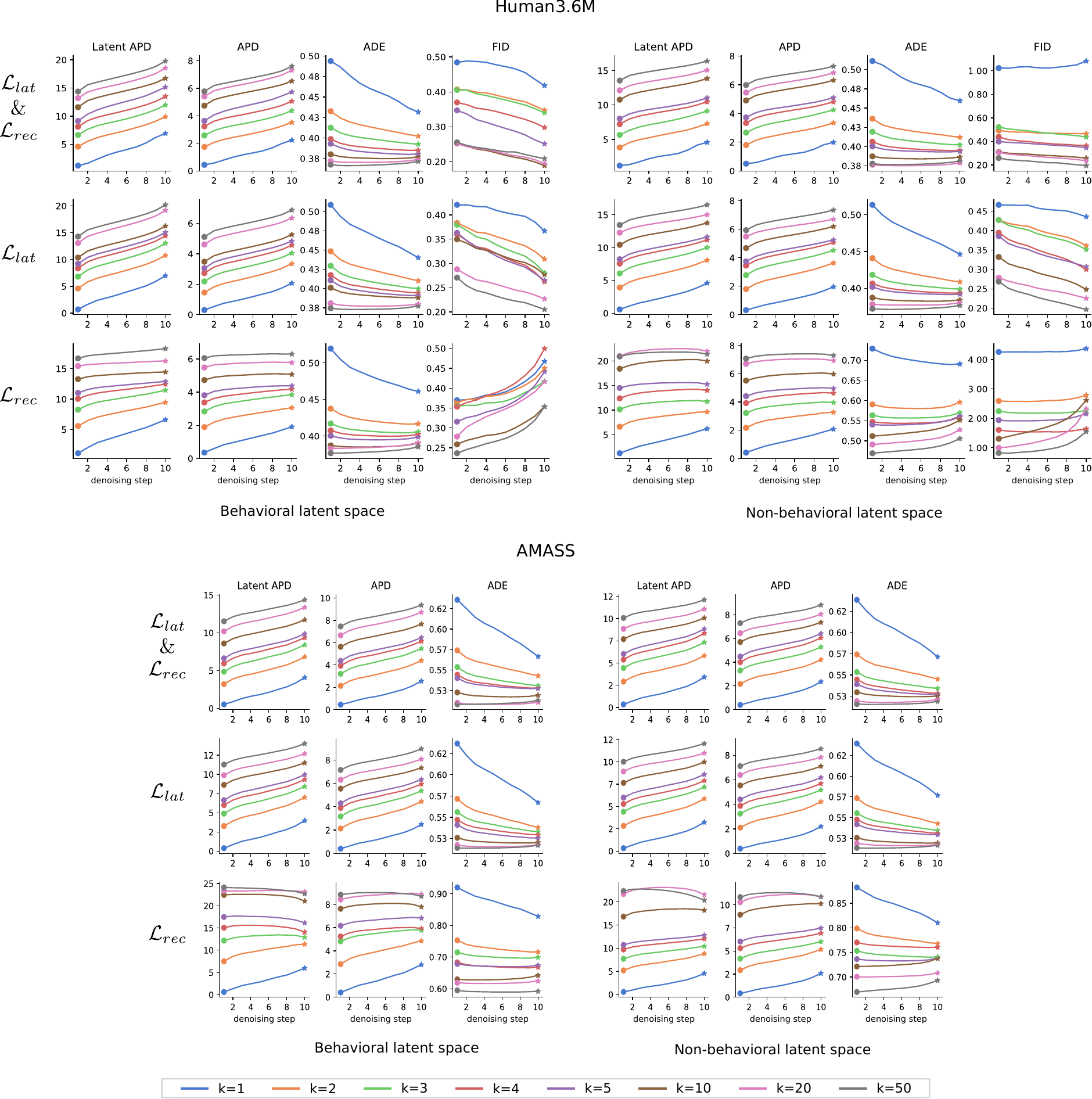}
    \caption{\textbf{Implicit diversity. }By increasing the value of $k$, the diversity is implicitly promoted in both the latent and reconstructed spaces (Latent APD, and APD). 
    We observe that this effect is not particular to the loss choice ($\mathcal{L}_{lat}$, $\mathcal{L}_{rec}$, or both) or the latent space construction (behavioral or not). Using the LDM to reverse the whole Markov chain of 10 steps (x-axis) helps improve diversity (APD), accuracy (ADE), and realism (FID) in general. Note that for $k>5$, only the diversity and the realism are further improved, and a single denoising step becomes enough to generate the most accurate predictions.}
    \label{fig:supp_ext_k_analysis}
\end{figure*}

We already showed that by increasing $k$, the diversity (APD), accuracy (ADE, FDE), and realism (FID) of \modelname{} improves. In fact, for large $k$ ($>5$), a single denoising step becomes enough to achieve state-of-the-art accuracy. Still, going through the whole reverse Markov diffusion chain helps the predicted behavior code move closer to the latent space manifold, thus generating more realistic predictions. In \autoref{fig:supp_ext_k_analysis}, we include the same analysis for all the models in the ablation study of the main paper. The results prove that the implicit diversity effect is not exclusive of either \modelname{}'s loss or behavioral latent space. %

\subsection{Inference times}
\label{subsec:supp_implementation_times}

We computed the time it takes \modelname{} and the state-of-the-art models to generate 50 samples for a single prediction on a single GTX 1080Ti GPU. We averaged the values across 50 runs of 100 sequences.
BeLFusion (320/323ms for H36M/AMASS) is slower than BeGAN (17/20ms), TPK (30/38ms), DLow(34/43ms), GSPS(5/7ms), and DivSamp (6/9ms).
Despite BeLFusion's slower inference (discussed as limitation in Sec.~5,), its $z_0$ parametrization allows it to be early-stopped and run, if needed, in real-time (BeLFusion\_D, 48/XXms) with similar accuracy in return for less diversity and worse APDE, FID, and CMD.

\section{Examples \textit{in motion}}
\label{sec:supp_visual_examples}

For each dataset, we include several videos where 10 predictions of \modelname{} are compared to those of methods showing competitive performance for H36M: TPK~\cite{walker2017theposeknows}, DLow~\cite{yuan2020dlow}, GSPS~\cite{mao2021gsps}, and DivSamp~\cite{dang2022diverse}. Videos are identified as `[dataset]\_[sample\_id]\_[class/subdataset]'. For example, `A\_6674\_GRAB' is sample 6674, which is part of the GRAB~\cite{taheri2020grab} dataset within AMASS (prefix `A\_'), and `H\_1246\_Sitting' is the sample 1246, which is part of a `Sitting' sequence of H36M (prefix `H\_'). The \textit{Context} column shows the observed sequence and freezes at the last observed pose. The \textit{GT} column shows the ground truth motion. %

In this section, we discuss the visual results by highlighting the main advantages provided by \modelname{} and showing some failure examples.

\textbf{Realistic transitioning. } By means of the behavior coupler, \modelname{} is able to transfer predicted behaviors to any ongoing motion with high realism. This is supported quantitatively by the FID and CMD metrics, and perceptually by our qualitative assessment (Sec. 4.3). Now, we assess it by visually inspecting several examples. For example, when the observation shows an ongoing fast motion (`H\_608\_Walking', `H\_1928\_Eating' or `H\_2103\_Photo'), \modelname{} is the only model that consistently generates a coherent transition between the observation and the predicted behavior. Other methods mostly predict a sudden stop of the previous action. This is also appreciated in the cross-dataset evaluation. For example, although the observation window of the `A\_103\_Transitions' clearly showcases a fast rotational dancing step, none of the state-of-the-art methods are able to generate a plausible continuation of the observed motion, and all of their predictions abruptly stop rotating. \modelname{} is the only method that generates predictions that slowly decrease the ongoing motion's rotational momentum to start performing a different action. A similar effect is observed in `A\_2545\_DanceDB', and `A\_10929\_HUMAN4D'.

\textbf{Context-driven prediction. } \modelname{}'s state-of-the-art APDE and CMD metrics show its superior ability to adjust both the \textit{motion speed} and \textit{motion determinacy} to the observed context. This results in sets of predictions that are, overall, more coherent with respect to the observed context. For example, whereas for `H\_4\_Sitting' \modelname{}'s predicted motions showcase a high variety of arms-related actions, its predictions for sequences where the arms are used in an ongoing action (`H\_402\_Smoking', `H\_446\_Smoking', and `H\_541\_Phoning') have a more limited variety of arms motion. In contrast, predictions from state-of-the-art methods do not have such behavioral consistency with respect to the observed motion. This is more evident in diversity-promoting methods like DLow, GSPS, and DivSamp, where the motion predicted is usually implausible for a person that is smoking or making a phone call. Similarly, in `H\_962\_WalkTogether', our method predicts motions that are compatible with the ongoing action of walking next to someone, whereas other methods ignore such possibility. In AMASS, \modelname{}'s capability to adapt to the context is clearly depicted in sequences with low-range motion, or where motion is focused on particular parts of the body. For example, \modelname{} adapts the diversity of predictions to the `grabbing' action present in the GRAB dataset. While other methods predict coordinate-wise diverse inaccurate predictions, our model encourages diversity within the short spectrum of the plausible behaviors that can follow (see `A\_7667\_GRAB', `A\_7750\_GRAB', or `A\_9274\_GRAB'). In fact, in `A\_11074\_HUMAN4D' and `A\_12321\_SOMA', our model is the only able to anticipate the intention of laying down by detecting subtle cues inside the observation window (samples \#6 and \#8).  In general, \modelname{} provides good coverage of all plausible futures given the contextual setting. For example, in `H\_910\_SittingDown', and `H\_861\_SittingDown' our model's predictions contain as many different actions as all other methods, with no realism trade-off as for GSPS or DivSamp.

\textbf{Generalization to unseen contexts. } As a result of the two properties above (realistic transitioning and context-driven prediction), \modelname{} shows superior generalization to unseen situations. This is quantitatively supported by the big step forward in the results of the cross-dataset evaluation. Such generalization capabilities are especially perceptible in the DanceDB\footnote{Dance Motion Capture DB, \url{http://dancedb.cs.ucy.ac.cy}.} sequences, which include dance moves unseen at training time. For instance, `A\_2054\_DanceDB' shows how \modelname{} can predict, up to some extent, the correct continuation of a dance move, while other methods either almost freeze or simply predict an out-of-context movement. Similarly, `A\_2284\_DanceDB' and `A\_1899\_DanceDB' show how \modelname{} is able to detect that the dance moves involve keeping the arms arising while moving or rotating. In comparison, DLow, GSPS, and DivSamp simply predict other unrelated movements. TPK is only able to predict a few samples with fairly good continuations to the dance step. Also, in `A\_12391\_SOMA', \modelname{} is the only method able to infer how a very challenging repetitive stretching movement will follow.

We also include some examples where our model fails to generate a coherent and plausible set of predictions. This mostly happens under aggressive domain shifts. For example, in `A\_1402\_DanceDB', the first-seen handstand behavior in the observation leads to \modelname{} generating several wrong movement continuations. Similarly to the other state-of-the-art methods, \modelname{} also struggles with modeling high-frequencies. For example, in `A\_1087\_DanceDB', the fast legs motion during the observation is not reflected in any prediction, although \modelname{} slightly shows it in samples \#4 and \#7. Even though less clearly, this is also observed in H36M. For example, in `H\_148\_WalkDog', none of the models is able to model the high-speed walking movement from the ground truth. Robustness against huge domain drifts and modeling of high-frequencies are interesting and challenging limitations that need to be addressed as future work.

\section{Qualitative assessment}
\label{sec:supp_mos}

\textbf{Selection criteria. } In order to ensure the assessment of a wide range of scenarios, we randomly sampled from three sampling pools per dataset. To generate them, we first ordered all test sequences according to the average joint displacement $D_i$ in the last 100 ms of observation. Then, we selected the pools by taking sequences with $D_i$ within 1) the top 10\% (high-speed transition), 2) 40-60\% (medium-speed transition), and 3) the bottom 10\% (low-speed transition). Then, 8 sequences were randomly sampled for each group. A total of 24 samples for each dataset were selected. These were randomly distributed in groups of 4 and used to generate 6 tests per dataset. Since each dataset has different joint configurations, we did not mix samples from both datasets in the same test to avoid confusion.

\begin{figure*}
    \centering
    \begin{minipage}{.48\textwidth}
        \centering
        \includegraphics[width=\textwidth]{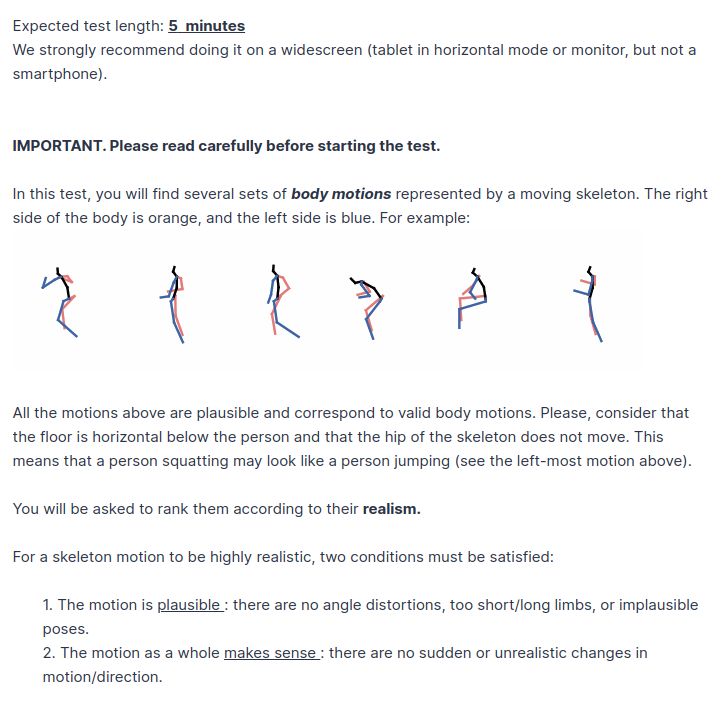}
        \label{fig:supp_mos_example}
    \end{minipage}%
    \hfill
    \begin{minipage}{0.48\textwidth}
        \centering
        \includegraphics[width=\textwidth]{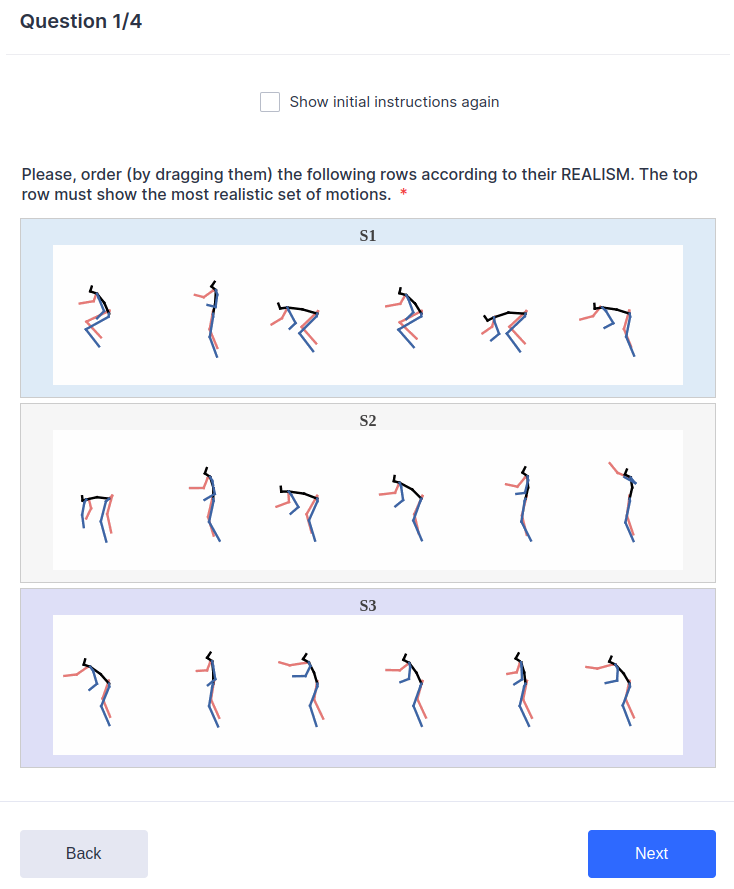}
        \label{fig:supp_mos_instructions}
    \end{minipage}
    \caption{\textbf{Questionnaire example. }On the left, instructions shown to the participant at the beginning. On the right, the interface for ranking the skeleton motions. All skeletons correspond to \textit{gif} images that repeatedly show the observation and prediction motion sequences. }
    \label{fig:supp_mos}
\end{figure*}

\setlength{\tabcolsep}{4pt}
\begin{table*}[t!]\renewcommand{\arraystretch}{0.9}
    \footnotesize
    \centering
    \begin{tabular}{l@{\hskip 8mm}cccc@{\hskip 8mm}cccc}
        \toprule
        & \multicolumn{4}{c}{Human3.6M\cite{ionescu2013h36m}} & \multicolumn{4}{c}{AMASS\cite{mahmood2019amass}} \\
        \midrule
        & Avg. rank & Ranked 1st & Ranked 2nd & Ranked 3rd & Avg. rank & Ranked 1st & Ranked 2nd & Ranked 3rd \\
        \midrule
        \multicolumn{9}{l}{Low-speed transition} \\
        \midrule
        GSPS & 2.238 $\pm$ 0.305 & 18.0\% & \textbf{40.4\%} & 41.6\% & 2.156 $\pm$ 0.595 & 22.6\% & \textbf{38.1\%} & 39.3\%\\
        DivSamp & 2.276 $\pm$ 0.459 & 15.7\% & 39.3\% & \textbf{44.9\%} & 2.210 $\pm$ 0.373 & 23.8\% & 31.0\% & \textbf{45.2\%}\\
        \modelname{} & \textbf{1.486 $\pm$ 0.225} & \textbf{66.3\%} & 20.2\% & 13.5\% & \textbf{1.634 $\pm$ 0.294} & \textbf{53.6\%} & 31.0\% & 15.5\%\\
        \midrule
        \multicolumn{9}{l}{Medium-speed transition} \\
        \midrule
        GSPS & 2.305 $\pm$ 0.466 & 13.8\% & \textbf{48.3\%} & 37.9\% & 2.025 $\pm$ 0.449 & 24.3\% & \textbf{50.0\%} & 25.7\%\\
        DivSamp & 2.396 $\pm$ 0.451 & 9.2\% & 36.8\% & \textbf{54.0\%} & 2.497 $\pm$ 0.390 & 10.8\% & 28.4\% & \textbf{60.8\%}\\
        \modelname{} & \textbf{1.299 $\pm$ 0.243} & \textbf{77.0\%} & 14.9\% & 8.0\% & \textbf{1.478 $\pm$ 0.424} & \textbf{64.9\%} & 21.6\% & 13.5\%\\
        \midrule
        \multicolumn{9}{l}{High-speed transition} \\
        \midrule
        GSPS & 2.194 $\pm$ 0.320 & 21.7\% & \textbf{40.2\%} & 38.0\% & 1.828 $\pm$ 0.468 & 44.9\% & 35.9\% & 19.2\%\\
        DivSamp & 2.345 $\pm$ 0.292 & 15.2\% & 32.6\% & \textbf{52.2\%} & 2.589 $\pm$ 0.409 & 6.4\% & 26.9\% & \textbf{66.7\%}\\
        \modelname{} & \textbf{1.461 $\pm$ 0.149} & \textbf{63.0\%} & 27.2\% & 9.8\% & \textbf{1.583 $\pm$ 0.287} & \textbf{48.7\%} & \textbf{37.2\%} & 14.1\%\\
        \midrule
        \multicolumn{9}{l}{All} \\
        \midrule
        GSPS & 2.246 $\pm$ 0.358 & 17.9\% & \textbf{42.9\%} & 39.2\% & 2.003 $\pm$ 0.505 & 30.5\% & \textbf{41.1\%} & 28.4\% \\
        DivSamp & 2.339 $\pm$ 0.393 & 13.4\% & 36.2\% & \textbf{50.4\%} & 2.432 $\pm$ 0.408 & 14.0\% & 28.8\% & \textbf{57.2\%}  \\
        \modelname{} & \textbf{1.415 $\pm$ 0.217} & \textbf{68.7\%} & 20.9\% & 10.4\% & \textbf{1.565 $\pm$ 0.332} & \textbf{55.5\%} & 30.1\% & 14.4\% \\
        \bottomrule
    \end{tabular}
    \caption{\textbf{Qualitative assessment.} \participants{} participants ranked sets of samples from GSPS, DivSamp, and \modelname{} by their realism. Lower average rank ($\pm$ std. dev.) is better.}
    \label{tab:supp_mos_results}\vspace{-0.3cm}
\end{table*}
\setlength{\tabcolsep}{6pt}

\textbf{Assessment details. } The tests were built with the \textit{JotForm}\footnote{\url{https://www.jotform.com/}} platform. Users accessed it through a link generated with \textit{NimbleLinks}\footnote{\url{https://www.nimblelinks.com/}}, which randomly redirected them to one of the tests. \autoref{fig:supp_mos} shows an example of the instructions and definition of realism shown to the user before starting the test (left), and an example of the interface that allowed the user to order the methods according to the realism showcased (right). Note that the instructions showed either AMASS or H36M ground truth samples, as both skeletons have a different number of joints. A total of \participants{} people answered the test, with 67 participating in the H36M study, and 59 participating in the AMASS one.

\textbf{Extended results.} Extended results for the qualitative study are shown in \autoref{tab:supp_mos_results}. We also show the results for each sampling pool, i.e., grouping sequences by the speed of the transition. The average rank was computed as the average of all samples' mean ranks, and the 1st/2nd/3rd position percentages as the number of times a sample was placed at 1st/2nd/3rd position over the total amount of samples available. We observe that the realism superiority of \modelname{} is particularly notable in the sequences with medium-speed transitions (77.0\% and 64.9\% ranked first in H36M and AMASS, respectively). We argue that this is partly promoted by the good capabilities of the behavior coupler to adapt the prediction to the movement speed and direction observed. This is also seen in the high-speed set (ranked third only in 9.8\% and 14.1\% of the cases), despite GSPS showing competitive performance on it.

\end{document}